\newcommand{\CLASSINPUTtoptextmargin}{0.75in}
\newcommand{\CLASSINPUTbottomtextmargin}{1.01in}
\documentclass[conference]{IEEEtran}
\usepackage{cite}
\usepackage{amsmath,amssymb,amsfonts}
\usepackage{algorithm}
\usepackage{algpseudocode}
\usepackage{graphicx}
\usepackage{textcomp}
\usepackage{booktabs}
\usepackage{tabularx}
\usepackage{enumitem}
\usepackage{xcolor}
\usepackage{multirow}
\usepackage{wrapfig}
\usepackage{subcaption}
\usepackage{soul}
\usepackage[moderate]{savetrees}
\usepackage{enumitem}
\setlist{nosep}
\setlist[itemize]{nosep}
\usepackage[font=footnotesize]{caption}

\newcommand{\names}{{\em COHORT }}
\newcommand{\namef}{{\em COHORT's\ }}

\begin{document}

\title{COHORT: Hybrid RL for Collaborative Large DNN Inference on Multi-Robot Systems Under Real-Time Constraints}

\author{
    \parbox{\linewidth}{\centering
        M. S. Anwar\textsuperscript{1}, A. Ravi\textsuperscript{1}, I. Ghosh\textsuperscript{1}, G. Shinde\textsuperscript{1}, C. Busart\textsuperscript{2}, N. Roy\textsuperscript{1} \\[0.5ex]
        \small \textsuperscript{1}University of Maryland Baltimore County \quad \textsuperscript{2}DEVCOM Army Research Laboratory \\[0.5ex]
        \small \{saeid.anwar, anuradha, gshinde1, indrajeetghosh, nroy\}@umbc.edu \quad carl.e.busart.civ@army.mil
    }
}
\maketitle

\begin{abstract} 
Large deep neural networks (DNNs), especially transformer-based and multimodal architectures, are computationally demanding and challenging to deploy on resource-constrained edge platforms like field robots. These challenges intensify in mission-critical scenarios (e.g., disaster response), where robots must collaborate under tight constraints on bandwidth, latency, and battery life, often without infrastructure or server support. To address these limitations, we present COHORT, a collaborative DNN inference and task-execution framework for multi-robot systems built on the Robotic Operating System (ROS). COHORT employs a hybrid offline–online reinforcement learning (RL) strategy to dynamically schedule and distribute DNN module execution across robots. Our key contributions are threefold: (a) Offline RL policy learning combined with Advantage-Weighted Regression (AWR), trained on auction-based task allocation data from heterogeneous DNN workloads across distributed robots, (b) Online policy adaptation via Multi-Agent PPO (MAPPO), initialized from the offline policy and fine-tuned in real time, and (c) comprehensive evaluation of COHORT on vision-language model (VLM) inference tasks such as CLIP and SAM, analyzing scalability with increasing robot/workload and robustness under . We benchmark COHORT against genetic algorithms and multiple RL baselines. Experimental results demonstrate that COHORT reduces battery consumption by 15.4\%, and increases GPU utilization by 51.67\%, while satisfying frame-rate and deadline constraints $2.55\times$ of the time.
\end{abstract}

\begin{IEEEkeywords}
Distributed Autonomous Systems, Hybrid RL, Advantage-Weighted Regression, MAPPO.
\end{IEEEkeywords}
\section{Introduction}

Modern search-and-rescue (SAR) operations increasingly employ autonomous robotic agents to support human responders in locating survivors and assessing hazardous environments~\cite{Chitikena2023-gj}. These robotic agents are required to interpret complex scenes, detect and track objects of interest, and respond to natural-language queries issued by human operators in real time. While individual agents are capable of navigating diverse terrains autonomously, collaborative operation among robots is essential~\cite{Hadidi2018Collaborative} to improve perception quality, navigation robustness, and system resilience—particularly in scenarios where one or more robots may fail during deployment. Collaboration further enables more effective utilization of limited onboard resources, which is critical because robotic agents are severely resource constrained and often operate for extended durations without access to recharging infrastructure.

Energy efficiency is a fundamental concern in SAR missions. Although each robotic agent is equipped with its own battery, uncoordinated task execution can rapidly deplete critical energy reserves, thereby reducing overall mission duration. For example, a Jackal unmanned ground vehicle (UGV) can operate on battery power for approximately 2–8 hours~\cite{McNulty2022-ft}, depending on the workload, with perception and navigation pipelines consuming the majority of available energy. While prior research has focused extensively on optimizing navigation efficiency, our work targets the optimization of perception workloads to extend mission lifetime.

Perception awareness is a cornerstone of SAR operations, enabling robots to identify humans in cluttered scenes, navigate debris-filled environments, and coordinate medical assistance with first responders. Recent advances in Vision–Language Models (VLMs) have significantly improved perception accuracy and generalization across diverse environments. While convolutional neural networks (CNNs) such as YOLOv8 have achieved notable success in object detection, they require extensive supervised training and domain-specific tuning. In contrast, VLMs generalize effectively across heterogeneous scenes, support open-ended reasoning, and respond directly to human queries—capabilities that are particularly valuable in SAR scenarios~\cite{chen2024taskclip}. However, these advantages come at a substantial computational cost: VLMs are resource-intensive~\cite{shinde2025survey}, demanding significant compute, memory, and energy resources to execute efficiently.

To mitigate these challenges, existing approaches often rely on the edge–cloud continuum to offload computation and reduce latency~\cite{Zhang2025VLM}. In disaster-response scenarios, however, reliable network connectivity to cloud or server infrastructure is often unavailable, rendering such approaches impractical. Consequently, robotic agents must operate in a fully decentralized manner, collaboratively sharing workloads and leveraging peers with surplus resources. A practical solution is to distribute computational workloads among robots and utilize agents with greater available resources to execute demanding tasks~\cite{9835126}. Several state-of-the-art studies~\cite{10413648, zhang2024edgeshard} have proposed optimization-based methods to partition and distribute DNN workloads across edge devices and cloud servers. Hongcai et al.~\cite{10473221} further propose a deep reinforcement learning framework to allocate computational tasks across multiple edge devices. However, these solutions are typically evaluated in simulation or controlled environments, do not account for real-world robotic deployments, and often overlook critical challenges related to scalability and resilience—particularly in cases where devices may dynamically join or leave the system.

Achieving effective collaborative inference in real-world robotic systems introduces several intertwined challenges. Heterogeneity is inherent: UGVs differ substantially in compute throughput, memory capacity, sensing modalities, and battery chemistry, rendering fixed or one-time model partitioning ineffective as operating conditions evolve. Resource volatility further complicates execution, as robot mobility and temperature variations cause available compute and energy budgets to fluctuate over time. Energy must also be treated as a first-class constraint: naïve scheduling decisions can prematurely drain critical robots, reducing mission coverage or forcing early withdrawal from operational zones.

To address these challenges, we propose \namef, a distributed large DNN inference framework that dynamically balances workloads across heterogeneous robotic agents based on real-time resource availability. Unlike prior work~\cite{11030760} that assumes homogeneous workloads or resources, \namef explicitly accounts for dynamic workloads and device heterogeneity. Moreover, existing frameworks such as auction-based scheduling~\cite{10839040} require frequent information exchange between agents to share resource states and bids—an impractical assumption in bandwidth-constrained SAR environments. In contrast, \namef emphasizes one-shot decision making and minimizes inter-agent communication.

Deep reinforcement learning frameworks such as Multi-Agent Proximal Policy Optimization (MAPPO)~\cite{Chao2022mappo} provide a powerful foundation for cooperative decision making in multi-agent systems. However, MAPPO typically requires a large number of online interactions to converge to stable policies. To mitigate this limitation, we introduce a hybrid reinforcement learning (HybridRL) strategy that combines offline and online learning. Specifically, \namef first collects offline execution traces using auction-based heuristic policies while executing heterogeneous DNN modules across multiple robots. These traces are used to train resource-specific agent policies and a centralized critic that captures inter-agent interactions. The resulting policies are then deployed on individual robots and fine-tuned online to adapt to changing workloads, varying team sizes, and dynamic environmental conditions. When new devices join the system, \namef initializes their policies by selecting pretrained agents with similar resource profiles—for example, deploying a Husky-trained agent on a laptop with comparable computational resources. In addition to efficient task allocation, \namef explicitly enforces real-time execution constraints, including application-specific frame-rate (FPS) requirements and execution deadlines. These constraints are integrated into the learning objective to ensure that collaborative inference remains both timely and energy-efficient.

We summarize our contributions as follows:

\noindent \textbf{1) Resource-Aware Distributed Large DNN Execution Framework:} \names introduces a two-stage reinforcement learning pipeline for distributed execution of large DNN workloads across heterogeneous robotic agents. The framework continuously monitors local and peer resource states and makes scheduling decisions with minimal inter-agent communication. By eliminating reliance on centralized servers, \names enables autonomous workload balancing under dynamic compute, network, and mobility conditions, improving robustness in real-world deployments.

\noindent \textbf{2) Fault-Tolerant Workload Reallocation Strategy:} We design a fault-tolerant mechanism that dynamically reallocates tasks in response to resource failures, compute overloads, or unexpected energy depletion. The framework maintains stability in perception accuracy, computational throughput, and battery health while preserving real-time responsiveness in uncertain and dynamic environments.

\noindent \textbf{3) Real-World Deployment and Evaluation:} We deploy \names on three heterogeneous robotic platforms—Husky, Jackal, and Spot—and evaluate its performance using real VLM workloads, including CLIP and SAM. Offline data collected using auction-based heuristics are used to train initial agent and critic policies, which are subsequently fine-tuned during live execution. We evaluate scalability by varying workloads and dynamically adding or removing robots, and demonstrate that \names achieves up to $1.9\times$ higher FPS, reduces battery consumption by $\approx$ 15\% and 41.27\% improved resource utilization compared to auction-based baselines.

\section{Background and Motivation}
We structure the related work into three categories: task execution across edge devices; distributed and collaborative inference via model partitioning and co-execution; and reinforcement learning–based resource management for adaptive offloading, partitioning, and scheduling under heterogeneous, time-varying conditions.
\subsection{Task Execution across Edge Devices}

Recent work on task execution across edge devices studies how to map heterogeneous, delay-sensitive workloads onto distributed compute resources while meeting strict accuracy and deadline constraints. In \cite{10188769}, on-demand DNN inference at an edge server with multiple early-exit networks is formulated as a joint service-selection and non-preemptive EDF scheduling problem, maximizing long-term utility while enforcing per-task accuracy and latency guarantees. In \cite{9998539}, execution-mode selection (local, edge, or drop) is cast as a supervised classification problem over observed network, workload, and battery features, achieving near-Lyapunov-optimal offloading decisions with very low decision latency. For vehicular settings, \cite{10061366} applies asynchronous actor–critic methods to coordinate task placement and multi-resource allocation across vehicles, roadside edge, and cloud, improving delay and energy under highly dynamic topologies.

To move beyond single decision makers and enable collaborative execution across multiple edge devices, several works formulate task execution as a multi-agent control problem. In \cite{10465255}, edge servers are modeled as partially observable agents in a Dec-POMDP, and multi-agent deep reinforcement learning is used to learn joint policies for task assignment and CPU allocation, improving users’ quality of experience without centralized coordination. The offline-to-online DRL framework in \cite{10473221} pre-trains offloading policies from historical traces and refines them online, yielding higher task success rates and robustness to non-stationary dynamics. From a robotics perspective, \cite{10613777} adopts sequential single-item auctions (and a connectivity-aware variant) to allocate non-atomic tasks across heterogeneous robots under communication constraints incrementally, reducing mission makespan while maintaining connectivity. Collectively, these approaches show that effective task execution across edge devices increasingly relies on intelligent, distributed decision-making, but most still treat each task as a monolithic unit, in contrast to our fine-grained, stage-level scheduling of complex perception pipelines across collaborating robots.

\subsection{Distributed and Collaborative Inference}
Recently, advancements in distributed and collaborative inference from adaptive model partitioning and device-edge co-inference to dependence-aware scheduling and over-the-air aggregation have markedly improved the accuracy-latency-efficiency trade-off for real-time perception \cite{11030760,10543048,10621314,10400970}. In \cite{liu2024ed,kim2022task}, Vision Transformers are adapted for distributed inference ED-ViT partitions prunes class-specialized sub-models for edge deployment, while TAViT uses client-side task-specific heads/tails with a shared server-side Transformer body and alternating training for multi-task learning without data sharing; \cite{ghasemi2024edgecloudai,mudvari2024splitllm} tackle collaborative edge–cloud compute—EdgeCloudAI filters on-edge with CNNs and offloads only cropped/downsampled key frames for cloud VLM analysis, and SplitLLM optimally splits LLMs across client/server via dynamic programming to balance compute and communication for higher throughput; \cite{you2024v2x} fuses vehicle/infrastructure cameras with textual prompts in a large VLM and contrastive learning for robust trajectory planning; \cite{patel2024splitwise} phase-splits LLM inference running compute-heavy prompt processing and memory-bound token generation on separate, better-suited machines with fast KV-cache transfer; and \cite{zhang2024edgeshard} shards LLM layers across heterogeneous edge devices and cloud with DP-based joint device selection and model partitioning to cut latency and boost throughput. Additionally, vehicular and industrial settings show similar gains: SAFE schedules collaborative DNN inference online in V2X to adapt partitioning/placement under fluctuating bandwidth/compute and reduce end-to-end latency \cite{10580543}, while joint deployment–offloading–resource allocation in IIoT optimally splits models between devices and edge to minimize delay under constraints \cite{9835126}.

\subsection{Reinforcement Learning for Resource Management}
Recently, reinforcement learning has been increasingly integrated into edge/IoT inference to adaptively partition DNNs, offload tasks, and schedule resources under non-stationary loads spanning RL-aided device–edge collaboration for cumulative delay minimization~\cite{10137743}, dynamic partitioning for just-in-time edge AI~\cite{11073623}, Lyapunov guided diffusion-DRL for joint partitioning/offloading in vehicular networks~\cite{10736570}, and queue-aware online DRL for IoT offloading stability~\cite{10253677}; complementary strands cover DRL-based resource management for industrial IoT inference~\cite{9384272}, collaborative model selection on heterogeneous edges~\cite{10697311}, DRL-driven resource allocation in cloud gaming via edge computing~\cite{9953046}, and collective DRL for distributed intelligence sharing at the edge \cite{9861735}. In~\cite{yang2023cooperative}, a priority-driven multi-agent deep reinforcement learning framework allows edge servers to make local offloading decisions while coordinating through a VRNN-based global state sharing model, enabling cooperative migration and dynamic CPU allocation to reduce completion time and improve resource utilization. ~\cite{li2023dnn} addresses vehicular edge computing by selecting task-specific DNN partition points, compressing intermediate features with quantization, and optimizing offloading with an improved particle swarm–genetic algorithm featuring time-varying parameters and adaptive crossover and mutation to minimize end-to-end delay under resource constraints. Rashid et.al. ~\cite{rashid2020monotonic} introduces QMIX, a centralized-training, decentralized-execution method that combines individual agent value functions via a monotonic nonlinear mixing network, with hypernetworks conditioned on the global state, to maintain consistency between centralized learning and decentralized policies. 

Vision–language models (VLMs), such as CLIP~\cite{radford2021learning}, BLIP-2~\cite{li2023blip2}, and Moondream, have demonstrated strong multimodal perception capabilities, including object detection, segmentation, and semantic reasoning. However, their high GPU throughput and memory demands exceed the capabilities of compact, power-constrained robotic platforms. Executing such models on a single robot rapidly depletes onboard resources, while cloud-based offloading is impractical in contested or disaster-response environments with limited connectivity~\cite{zhou2019edgeai}. Moreover, existing frameworks largely overlook scalability and resilience under real-world operating conditions with strict real-time constraints. To address these limitations, we propose \namef, a resource-aware, scalable, and resilient task-execution framework that enables collaborative deployment of large DNN models on distributed robotic agents in network-constrained environments.

\section{\names Framework}
We present \names, an end-to-end framework for prompt-driven collaborative perception and resource-aware computational task execution in heterogeneous multi-robot systems. \names employs a reinforcement-learning (RL) based decision-making framework that decides whether to execute the computational task on device or offload to a nearby peer-device (known as target). 

\subsection{System Architecture and Input}
\begin{figure}[!ht]
   \includegraphics[width=\columnwidth]{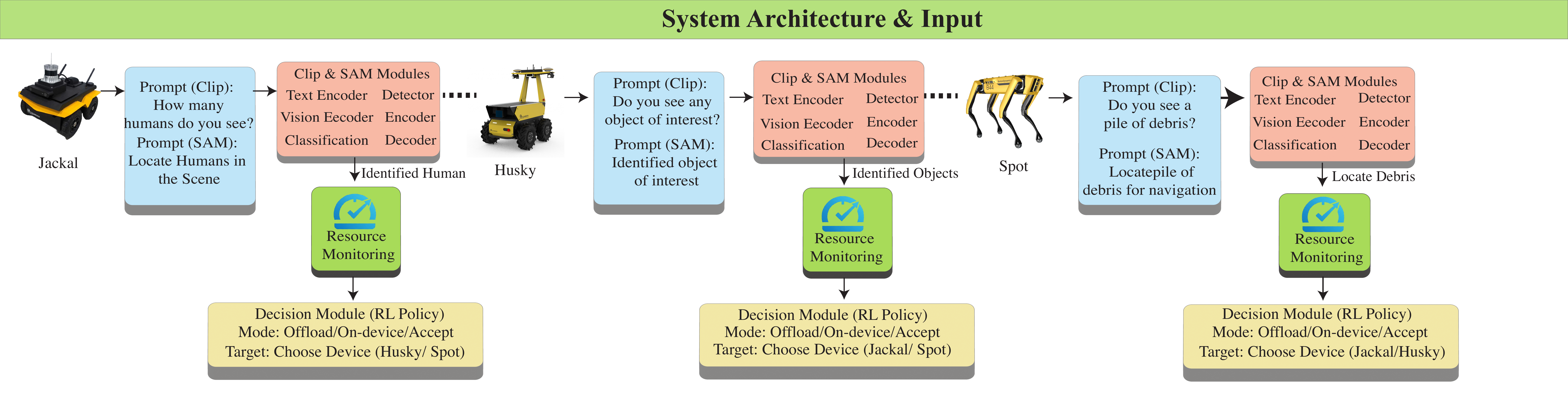}
    \caption{\names System Architecture}
    \label{fig:system_architecture}
\end{figure}

As illustrated in Fig.~\ref{fig:system_architecture}, the system consists of multiple heterogeneous robotic agents, including \textit{Jackal}, \textit{Husky}, and \textit{Spot}, each operating as an autonomous peer. Each robot receives prompts that specify perception objectives (e.g., human detection, object identification, or debris localization). These prompts are processed by paired CLIP and SAM modules that jointly perform semantic understanding and spatial grounding. The CLIP and SAM pipelines are pre-partitioned into six modular execution units: (a) detector, encoder, and decoder for SAM, and (b) text encoder, vision encoder, and classification for CLIP—supporting flexible scheduling and resource-aware offloading.

The perception pipeline outputs task-specific semantic results (e.g., identified humans, objects of interest, or obstacle locations) and is coupled with continuous resource monitoring (CPU and GPU utilization, memory usage, battery state of charge, and task queue) on each robot. These resources are observed by a local decision module, governed by an RL policy, which decides whether to offload a task or execute it on the device. The RL policy is trained to balance resource consumption while meeting real-time constraints, such as FPS and latency (deadline). The decision module determines the execution mode for each task: \textit{on-device execution}, \textit{offloading to a peer}, or \textit{accepting an incoming workload}. When offloading is selected, the policy also selects a target peer (e.g., Husky, Spot, or Jackal), enabling flexible, adaptive collaboration across heterogeneous platforms.

\subsection{Offline Training Pipeline}
\begin{figure}[!ht]
    \centering
   \includegraphics[width=.49\textwidth, ]{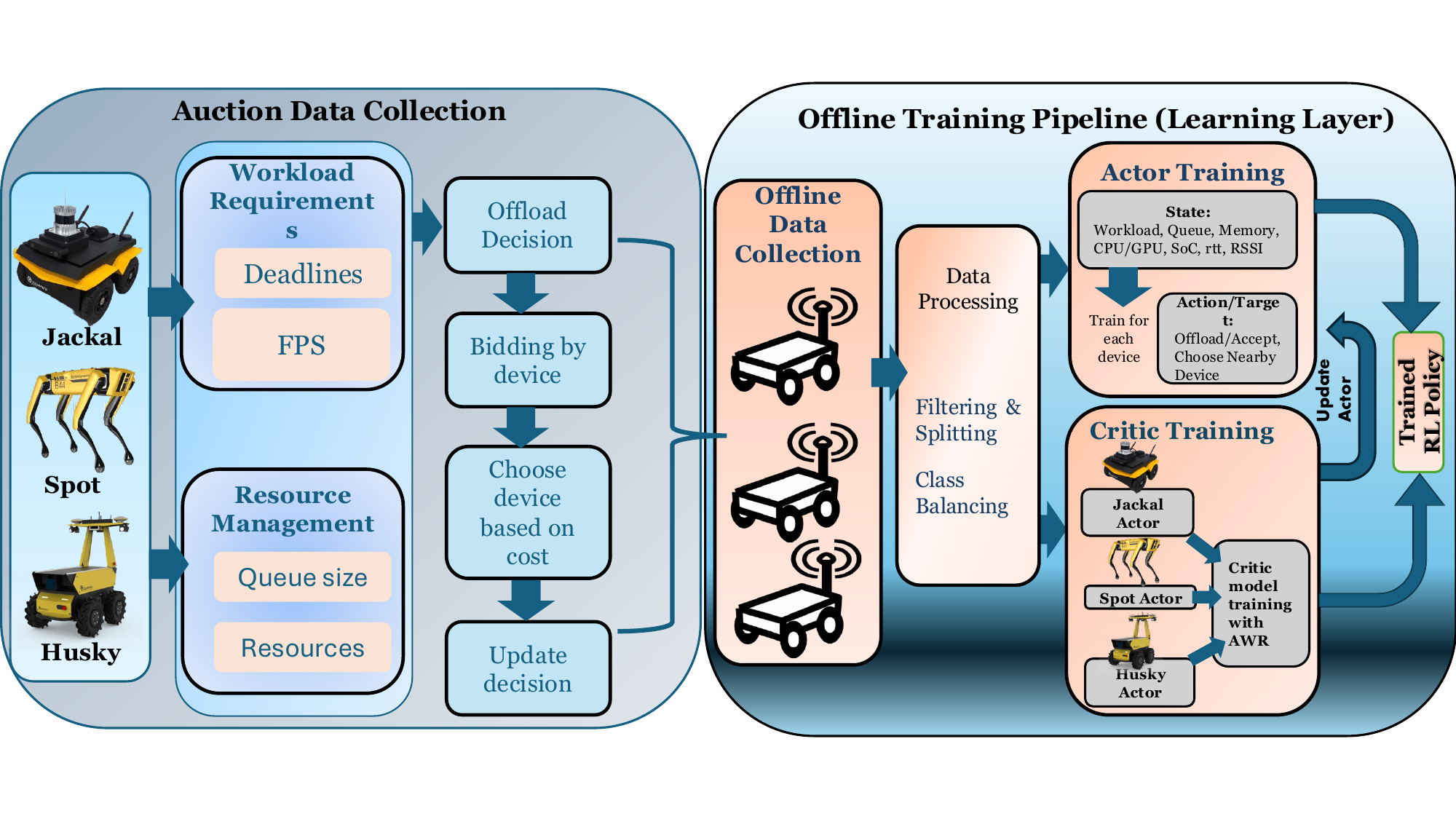}
    \caption{\names Offline RL Training}
    \label{fig:offline}
\end{figure}

Figure~\ref{fig:offline} illustrates the offline training pipeline used to learn the RL-based offloading policy. During data collection, all robots participate in an auction-based coordination process. For each task, workload requirements such as target frame rate and deadline constraints are combined with real-time resource states, including queue lengths and available compute capacity. Robotic agents submit bids reflecting their estimated execution cost, and task assignment decisions are recorded along with execution outcomes. The collected data from all agents are aggregated into an offline dataset and undergo preprocessing, including filtering, splitting, and class balancing. Policy learning follows an actor--critic paradigm. The actor network is trained to map observed system states (workload characteristics, resource availability, network latency, and battery status (SoC)) to offloading actions and target selection decisions. The critic network is trained using Advantage Weighted Regression (AWR) to evaluate action quality and provide stable learning signals. This offline phase yields a trained offline RL policy that captures efficient offloading strategies across diverse operating conditions (varying workloads, fps, and deadlines, battery SoC) on heterogeneous robots.

\subsection{Online Policy Execution}
\begin{figure}[!ht]
    \centering
    \includegraphics[width=.50\textwidth]{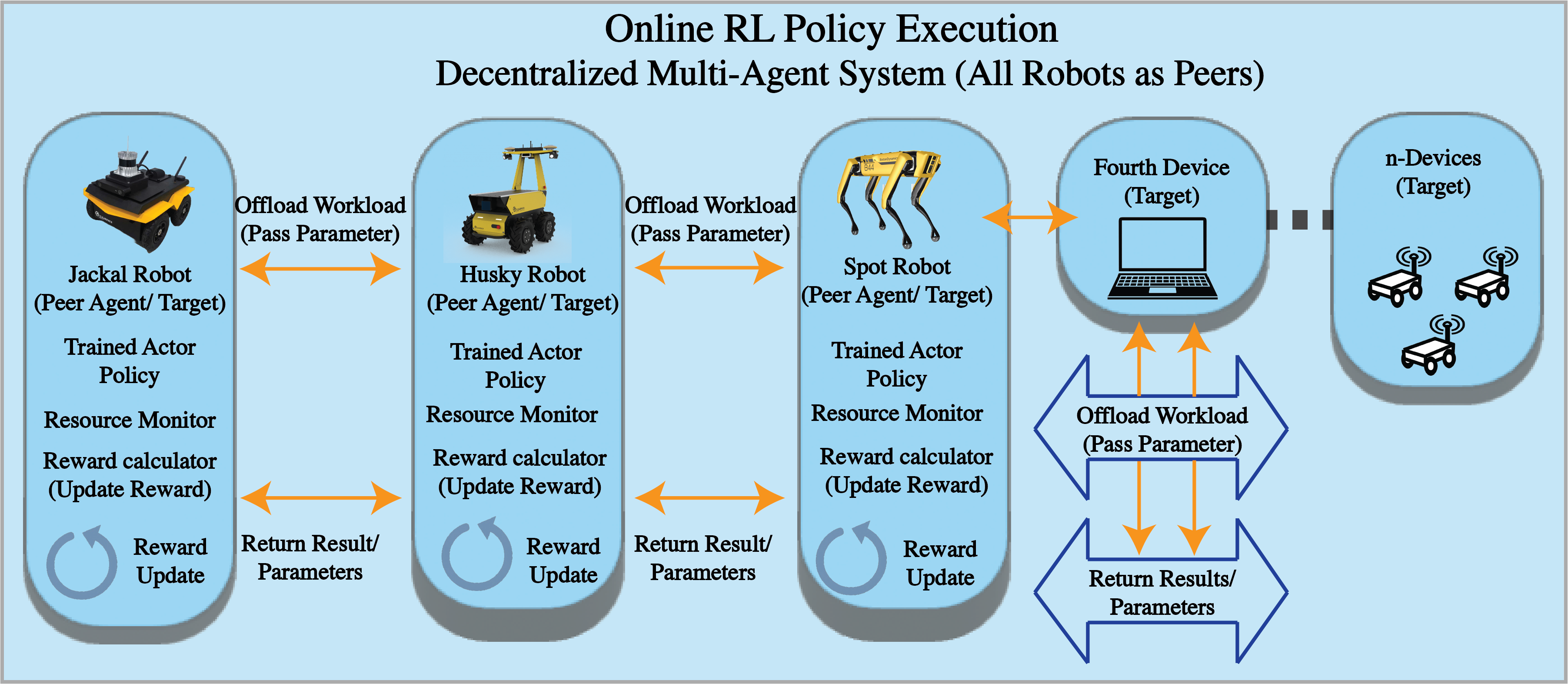}

    \caption{\names RL online Training }
    \label{fig:online}
\end{figure}

During online execution (Fig.~\ref{fig:online}), the trained RL policy is deployed in a fully decentralized multi-agent setting, where all robots operate as peers without centralized control. Each robot independently executes its trained actor policy, continuously monitoring local resources and evaluating incoming or generated tasks. When a task is offloaded, workload parameters are transmitted to the selected peer, which executes the task and returns results to the requesting agent. Throughout execution, each robot computes reward signals based on task latency, deadline satisfaction, energy consumption, and resource utilization. These rewards enable online performance tracking and can be used to support continued policy refinement. This decentralized execution framework employs an on-policy multi-agent reinforcement learning (MAPPO) formulation, where policies are updated using reward signals from fresh, real-time rollouts during execution. By continuously incorporating feedback on deadline violations, FPS targets, energy, and resources, the system adapts to dynamic workloads and resource conditions while maintaining efficient task execution.

\section{Problem formulation}

In this section, we formally introduce \names framework, which comprises (a) offline data collection using an auction-based cooperative scheduler, (b) offline RL policy training for each participating agent, and (c) online RL policy fine-tuning. We first model the distributed offloading decision policy as a cooperative partially observable stochastic game (POSG) to collect extensive data under varying operating conditions. In this stage, each agent participates in the auction to choose the best cooperative policy. We then describe a three-phase curriculum—behavior cloning (Phase~A) under centralized training and decentralized execution (CTDE), where a shared MAPPO actor replaces the hand-crafted auction cost (described by POSG) and selects the winner of each stage auction, offline advantage-weighted regression (Phase~B), and online MAPPO fine-tuning with Lagrangian constraints (Phase~C)—that enables safe, sample-efficient deployment on real Husky, Jackal, and Spot robots while meeting strict latency and FPS targets.

\subsection{Data Collection: Auction-Based Cooperative Decision Making}

To collect data for offline RL policy training, we cast offloading decision policy as a cooperative partially observable stochastic game (POSG) $\langle \mathcal{I},\mathcal{S},\{\mathcal{O}^i\}_{i\in\mathcal{I}},
\{\mathcal{A}^i\}_{i\in\mathcal{I}},\mathcal{T},r,\gamma\rangle$ 
Let $\mathcal{I}=\{1,\dots,N\}$ denote the $N$ robotic agents. We consider a fixed $L$-stage perception pipeline (e.g., \texttt{samA}$\!\rightarrow\!\cdots\!\rightarrow$\texttt{clipC}); processing a single RGB frame induces a finite-horizon trajectory of length $L$ with timesteps $t=0,\ldots,L{-}1$ corresponding to successive stages.


\paragraph{Observations.}
At the start of timestep $t$, agent $i\!\in\!\mathcal{I}$ receives
\[
    o_t^i = [\,r_t^i,\ n_t^i,\ c_t\,],
\]
where
\[
    r_t^i =
    [\,\text{battery\_horizon},\ \text{SoC},\ \text{power},\ \text{temp},\ 
      \text{cpu},\ \text{gpu},\ \text{ram},\ \text{queue}\,]_t^i,
\]
\[
    n_t^i = [\,\text{RSSI},\ \text{RTT}\,]_t^i,
\]
and
\[
    c_t =
    [\,\mathrm{onehot}(\text{stage}_t),\ \text{deadline\_ms}_t,\ \text{tensor\_bytes}_t\,].
\]
Optionally, a one-hot or learned embedding of the robot identity can be concatenated
to $o_t^i$ to allow specialization under shared parameters.
Each robot observes only its local telemetry $r_t^i,n_t^i$ and the shared stage
context $c_t$.

\paragraph{Actions and auction mechanism.}
At each stage $t$, every robot $i$ submits a continuous bid
$a_t^i\in\mathbb{R}$ that serves as a learned cost proxy for executing the 
current stage.
Let $\mathcal{I}_t\subseteq\mathcal{I}$ be the subset of robots that respond
within the bid window, and define the availability mask
$m_t^i=\mathbb{I}[i\in\mathcal{I}_t]$.
The executor is chosen by a deterministic auction rule
\[
    w_t=\arg\min_{i\in\mathcal{I}_t} a_t^i.
\]
The winner $w_t$ executes the stage (locally if it is the host; otherwise as an
offload target). This auction couples the agents' actions through the shared
winner index $w_t$.

\paragraph{Dynamics.}
Let $s_t\in\mathcal{S}$ denote the latent global state (resources and network
conditions across all robots).
After the winning robot executes the stage, the system evolves according to
\[
    s_{t+1}\sim\mathcal{T}\!\left(
        s_{t+1}\mid s_t,\ a_t^1,\ldots,a_t^N
    \right),
\]
capturing the effects on compute load, energy, queue lengths, and connectivity.

\paragraph{Action bounds (implementation).}
For numerical stability, bids are clipped to a bounded interval
$a_t^i\in[a_{\min},a_{\max}]$ (e.g., $[0,400]$) before transmission.
All continuous features in $o_t^i$ and the shaped reward $r_t$ are normalized,
and entropy regularization is used to encourage sufficient exploration.

\subsection{Three-Phase Training}

Training reinforcement learning (RL) agents directly on physical robotic
hardware is challenging due to sample inefficiency, safety risks, and the risk
of catastrophic battery drain during early exploration.
We address these challenges with a three-phase curriculum that transitions from
supervised imitation of the existing auction heuristic (Phase~A), to offline
policy improvement (Phase~B), and finally to online MAPPO fine-tuning with
CTDE (Phase~C).
Across all three phases we use \emph{shared} actor--critic weights for all
robots, with robot identity optionally encoded in the input.

\subsubsection{Phase A: Behavior Cloning (BC) for Safe Initialization}

Phase~A addresses the cold-start problem by initializing the shared policy
$\pi_\theta$ to imitate a stable decentralized auction-based heuristic.
This yields a baseline performance guarantee and prevents random, unsafe
exploration when the RL policy is first deployed.

We collect offline logs from the real system across all robots,
\[
    \mathcal{D} = \{(o_t, a_t, m_t, \text{robot\_id}_t)\},
\]
where $o_t$ is the local observation used at decision time, $a_t$ is the action
taken by the heuristic (e.g., bidding mode, offloading target), and $m_t$
encodes binary masks that enforce feasibility (e.g., prevent offloading to a
disconnected peer).
A shared multi-head actor network (Mode, Target, Capacity) is trained to
minimize a masked cross-entropy loss against the heuristic decisions:
\begin{equation}
    \mathcal{L}_{\mathrm{BC}}(\theta)
    =
    -\mathbb{E}_{(o_t, a_t, m_t) \sim \mathcal{D}}
    \left[
        \sum_{k \in \mathcal{K}}
        \mathbb{I}(m_{t,k})\,
        \log \pi_\theta(a_{t,k} \mid o_t)
    \right],
\end{equation}
where $\mathcal{K}$ indexes the policy heads and $\mathbb{I}(m_{t,k})$ is $1$
only if branch $k$ is valid according to the mask. Because the same $\theta$ is shared across robots, Phase~A produces a single policy that reproduces the baseline heuristic behavior on Husky, Jackal, and Spot while respecting connectivity and capacity constraints.

\paragraph{Reward and constraints.}
All robots share a scalar team reward
\begin{equation}
    r_t
    =
    \mathrm{FPS}_t
    \;-\;
    \lambda_D\,\mathbb{I}[\text{deadline\_miss}_t]
    \;-\;
    \lambda_E\,\mathrm{EnergyJ}_t,
\end{equation}
where $\mathrm{FPS}_t$ is the application-level frame rate at step $t$,
$\mathrm{EnergyJ}_t$ is the incremental energy consumption, and
$\lambda_D,\lambda_E>0$ control the trade-off between performance and battery
usage.
In practice, $r_t$ is \emph{shaped} using four measured signals:
(i) deadline slack (relative to the stage deadline),
(ii) end-to-end RTT to the offload target,
(iii) per-stage processing time, and
(iv) offloading time (network transfer duration).
These shaping terms provide dense feedback beyond binary deadline misses,
accelerating learning and reducing unsafe early exploration on the physical
robots.
Trajectories terminate after the final stage $t=L{-}1$ or upon failure.

\paragraph{Objective.}
We learn a shared stochastic policy $\pi_\theta$ with parameter sharing across
agents, where each robot draws its action from
$\pi_\theta(a_t^i\!\mid o_t^i)$ given its local observation.
The team objective is
\begin{equation}
    \max_\theta\ 
    \mathbb{E}\!\left[
        \sum_{t=0}^{L-1}\gamma^t r_t
    \right],
    \quad \gamma\in(0,1).
\end{equation}
This induces cooperative behavior where each robot's bidding strategy is aligned
with the global throughput/latency/energy trade-off.

\paragraph{CTDE value function (MAPPO).}
During training we adopt a MAPPO-style CTDE formulation with a centralized critic
$V_\phi$.
We construct a masked global state
\begin{equation}
    s_t^{\mathrm{CT}}
    =
    \big[
        \tilde{o}_t^1\!\odot\! m_t^1,\ 
        \ldots,\ 
        \tilde{o}_t^N\!\odot\! m_t^N,\ 
        c_t
    \big],
\end{equation}
where $\tilde{o}_t^i$ is the normalized observation of robot $i$ and
unavailable robots are zero-filled and masked by $m_t^i$.
The centralized critic $V_\phi(s_t^{\mathrm{CT}})$ is shared across all
robots and is only used during training; at execution time, each robot applies
the same shared policy $\pi_\theta(\cdot\mid o_t^i)$ to its own local
observation $o_t^i$.

\paragraph{Winner coupling and masks.}
The auction couples the agents through $w_t$, but MAPPO handles this coupling
by optimizing with the shared team reward $r_t$, the centralized critic
$V_\phi(s_t^{\mathrm{CT}})$, and the availability masks $\{m_t^i\}_{i=1}^N$
in both the policy and value losses.
This allows the critic to reason about cross-robot interactions (e.g., which
robot should win a given stage) while keeping execution decentralized.

\subsubsection{Phase B: Offline Policy Improvement via Advantage-Weighted Regression}

Phase~B transitions from pure imitation to RL without requiring new real-world
interactions.
The goal is to bias the shared policy toward historical decisions that yielded
higher returns (e.g., lower latency, better offloading choices), effectively
filtering out suboptimal actions taken by the heuristic.
We employ Advantage-Weighted Regression (AWR) on the same multi-robot dataset.

First, we train a centralized critic $V_\phi$ on the CTDE state
$s_t^{\mathrm{CT}}$ constructed from logs:
\begin{equation}
    \mathcal{L}_{V}(\phi)
    =
    \mathbb{E}_{\tau \sim \mathcal{D}}
    \left[
        \sum_{t=0}^{T}
        \big(
            V_\phi(s_t^{\mathrm{CT}}) - \hat{R}_t
        \big)^2
    \right],
\end{equation}
where $\hat{R}_t$ is a bootstrapped return computed from the shaped reward,
which combines deadline slack, RTT, per-stage processing time, and measured
offloading time into a single scalar signal (plus energy and FPS terms as in
$r_t$).
Next, we estimate advantages using Generalized Advantage Estimation (GAE):
\begin{equation}
    A_t^{\text{GAE}}
    =
    \sum_{l=0}^{\infty} (\gamma \lambda)^l \delta_{t+l},
    \quad
    \delta_t = r_t + \gamma V_\phi(s_{t+1}^{\mathrm{CT}}) - V_\phi(s_t^{\mathrm{CT}}),
\end{equation}
where $\lambda$ is the GAE smoothing parameter.
The shared policy is then updated to up-weight actions that outperform the
critic's expectation:

\begin{IEEEeqnarray}{rCl}
\mathcal{L}_{\mathrm{AWR}}(\theta)
&=&
-\mathbb{E}_{(o_t,a_t)\sim \mathcal{D}}
\Bigl[
\exp\!\Bigl(
\frac{1}{\beta}\,\mathrm{clip}(A_t,-\beta,\beta)
\Bigr)\nonumber\\
&&\qquad\qquad\cdot \log \pi_\theta(a_t \mid o_t)
\Bigr].
\end{IEEEeqnarray}

where $\beta$ is a temperature hyperparameter controlling the strength of
filtering.
Because both actor and critic share parameters across robots, Phase~B yields a
\emph{single} warm-started policy that is typically strictly better than the
hand-crafted auction heuristic before any online interaction.

\begin{algorithm}[b]
\caption{Three-Phase Training: BC $\rightarrow$ AWR $\rightarrow$ MAPPO (On-policy)}
\label{alg:three_phase_training}
\begin{algorithmic}[1]
  \State \textbf{Inputs:} multi-robot logs $\mathcal{D}$, shared actor $\pi_\theta$, critic $V_\phi$

  \State \textbf{Phase A: Behavior Cloning (offline)}
  \For{mini-batches $(o_t,a_t,m_t)$ from $\mathcal{D}$}
    \State minimize $\mathcal{L}_{\mathrm{BC}}(\theta)$ using masked cross-entropy
  \EndFor

  \State \textbf{Phase B: AWR (offline improvement)}
  \State fit $V_\phi$ on centralized states $s_t^{\mathrm{CT}}$ and returns $\hat{R}_t$
  \For{mini-batches $(o_t,a_t)$ from $\mathcal{D}$}
    \State compute advantages $A_t$ via GAE
    \State minimize $\mathcal{L}_{\mathrm{AWR}}(\theta)$ to up-weight high-advantage actions
  \EndFor

  \State \textbf{Phase C: MAPPO On-policy}
  \Loop{over online roll-outs}
    \For{each stage decision $t$}
      \State each robot $i$ samples bid $a_t^i \sim \pi_\theta(\cdot \mid o_t^i)$
      \State winner $w_t = \arg\min_i a_t^i$ executes the stage (local or offload)
      \State observe shaped, penalized reward $r'_t$ and next state
    \EndFor
    \State update $V_\phi$ and $\pi_\theta$ with PPO (centralized state $s_t^{\mathrm{CT}}$)
    \State update dual variables $\lambda$ via projected sub-gradient ascent
  \EndLoop
\end{algorithmic}
\end{algorithm}

\subsubsection{Phase C: Online MAPPO with Lagrangian Constraints (CTDE)}

In Phase~C, the pre-trained agents interact with the live multi-robot system.
We employ a MAPPO-style on-policy update with CTDE: all robots share the same
actor $\pi_\theta$ and centralized critic $V_\phi$, but make decisions based on
their own local observations $o_t^i$.
Unlike the original auction baseline where the winner was determined
by a fixed cost, the continuous bids in Phase~C are produced by
the shared MAPPO actor, so the RL policy directly controls which robot wins each
auctioned stage.
The critic conditions on the masked global state $s_t^{\mathrm{CT}}$ as defined
above.

To handle hard energy and latency constraints, we use a Lagrangian relaxation.
We define a penalized reward
\begin{equation}
    r'_t
    =
    r_t
    -
    \sum_{i} \lambda_{E,i} P_i^{(t)}
    -
    \lambda_{D}\,\mathbb{I}(\text{Miss}_t),
\end{equation}
where $r_t$ is the shaped team reward (encoding deadline slack, RTT, processing
time, and offloading time), $P_i^{(t)}$ is the power consumption of robot $i$
at step $t$, and $\mathbb{I}(\text{Miss}_t)$ indicates a deadline miss.
Dual variables $\lambda_{E,i}$ and $\lambda_D$ are updated online to penalize
constraint violations.

The shared policy is updated using the PPO clipped surrogate objective:
\begin{equation}
    \mathcal{L}^{\mathrm{CLIP}}(\theta)
    =
    \mathbb{E}_t
    \left[
        \min\!\left(
            \rho_t(\theta)\,\hat{A}_t,\ 
            \mathrm{clip}(\rho_t(\theta), 1-\epsilon, 1+\epsilon)\,\hat{A}_t
        \right)
    \right],
\end{equation}
where
$\rho_t(\theta) = \frac{\pi_\theta(a_t\mid o_t)}{\pi_{\theta_{\text{old}}}(a_t\mid o_t)}$
and $\hat{A}_t$ is the advantage estimated from the penalized rewards $r'_t$
using the centralized critic.
The dual variables are updated via projected sub-gradient ascent; for a
constraint $C \leq C_{\text{limit}}$, we use
\begin{equation}
    \lambda
    \leftarrow
    \big[
        \lambda + \alpha_\lambda (C_{\text{measured}} - C_{\text{limit}})
    \big]_+,
\end{equation}
where $\alpha_\lambda$ is the dual learning rate and $[\cdot]_+=\max(0,\cdot)$.
This MAPPO+CTDE formulation allows us to fine-tune a \emph{single} shared
policy online, while still logging and analyzing per-robot actor/critic losses
and rewards by tagging each sample with its \texttt{robot\_id}.
As a result, the final policy maximizes throughput (FPS) under tight energy and
latency budgets across heterogeneous robots.


\begin{figure}[b]
    \centering
    \begin{subfigure}[b]{0.155\textwidth}
        \centering
        \includegraphics[width=\linewidth]{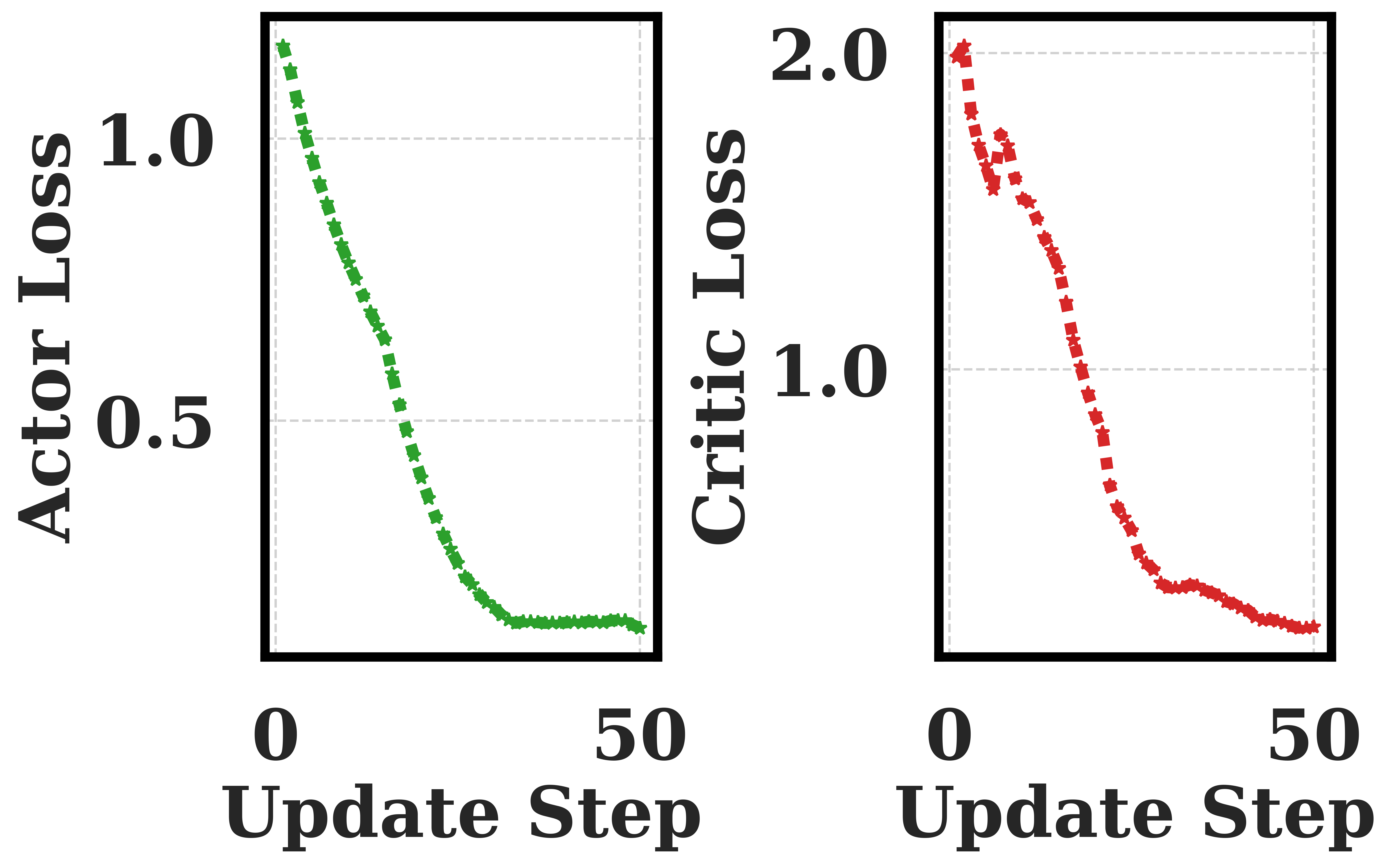}
        \subcaption{Husky}
        \label{fig:husky_loss}
    \end{subfigure}
    \hfill
    \begin{subfigure}[b]{0.155\textwidth}
        \centering
        \includegraphics[width=\linewidth]{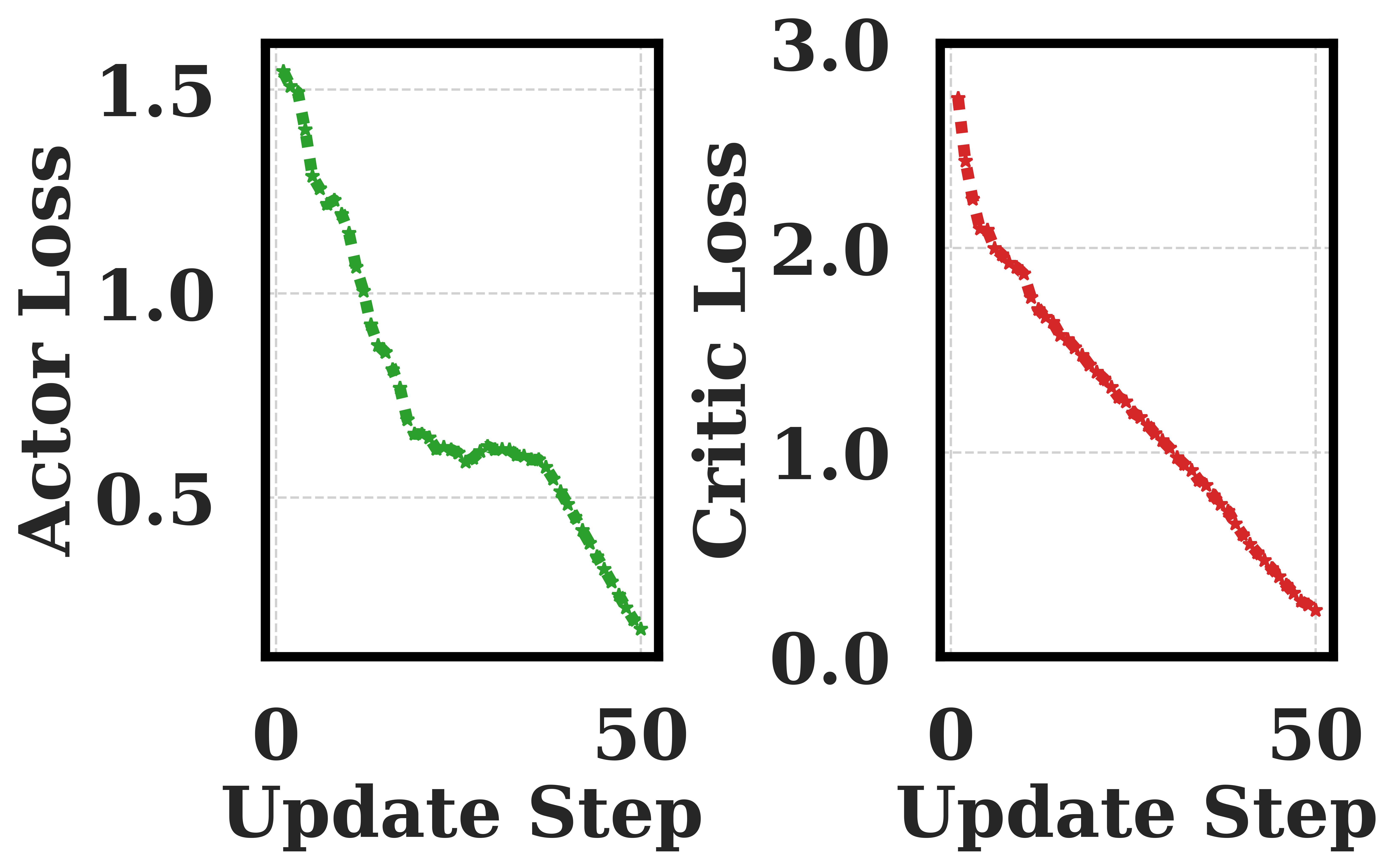}
        \subcaption{Jackal}
        \label{fig:jackal_loss}
    \end{subfigure}
    \hfill
    \begin{subfigure}[b]{0.155\textwidth}
        \centering
        \includegraphics[width=\linewidth]{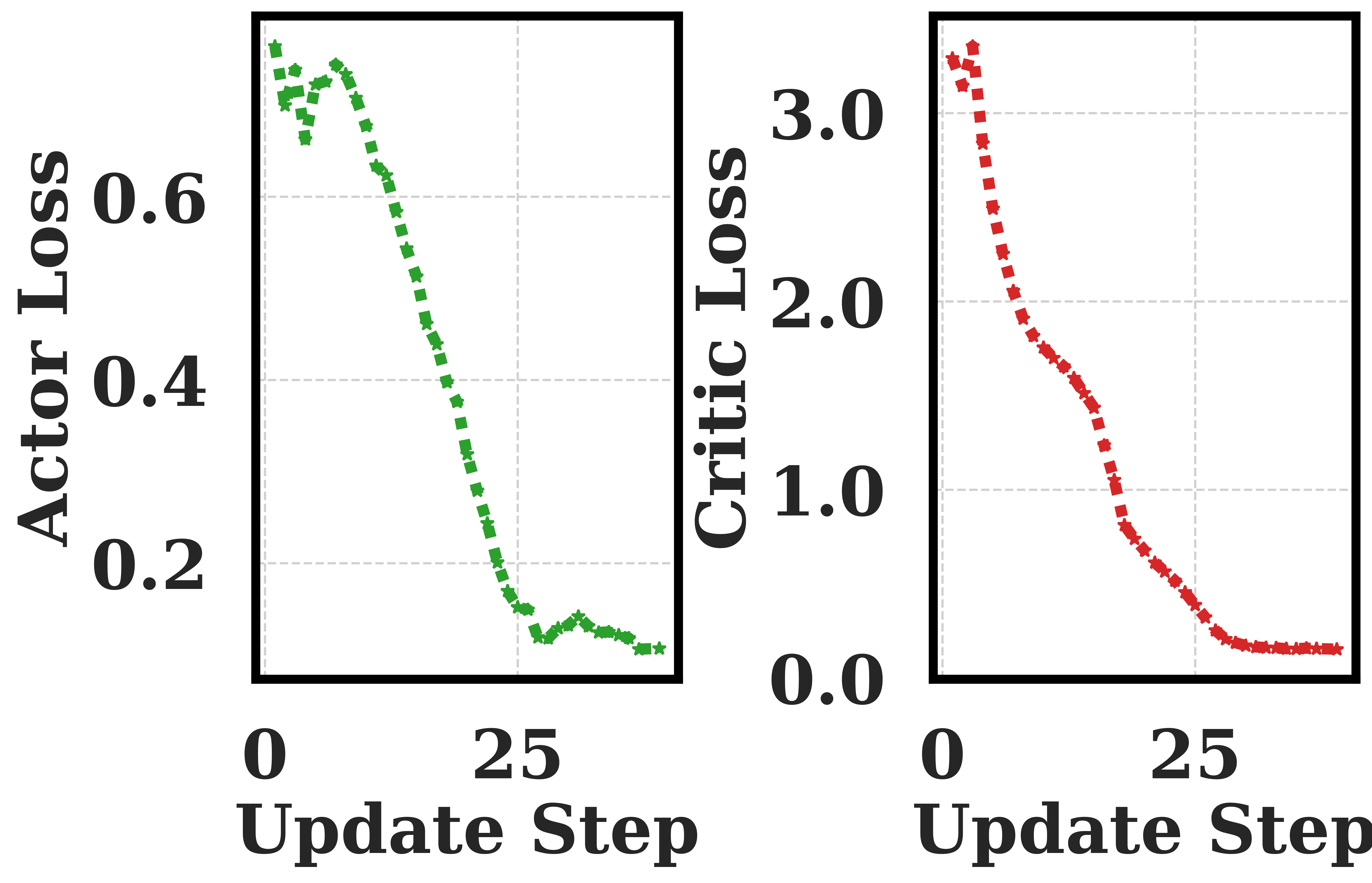}
        \subcaption{Spot}
        \label{fig:spot_loss}
    \end{subfigure}
    \caption{Actor and critic loss convergence across training updates for Husky, Jackal, and Spot.}
    \label{fig:robot_loss_comparison}
\end{figure}

The policy is updated using the PPO clipped surrogate objective to ensure stable  improvement:

\begin{equation}
    \mathcal{L}^{CLIP}(\theta) = \mathbb{E}_t \left[ \min \left( \rho_t(\theta) \hat{A}_t, \text{clip}(\rho_t(\theta), 1-\epsilon, 1+\epsilon) \hat{A}_t \right) \right]
\end{equation}

where $\rho_t(\theta) = \frac{\pi_\theta(a_t|o_t)}{\pi_{\theta_{\text{old}}}(a_t|o_t)}$ is the probability ratio, and $\hat{A}_t$ is the advantage estimated using the penalized rewards $r'_t$.

Simultaneously, the dual variables $\lambda$ are learned via sub-gradient ascent to adaptively penalize constraint violations. For a constraint $C \leq C_{\text{limit}}$, the multiplier update rule is:

\begin{equation}
    \lambda \leftarrow \left[ \lambda + \alpha_\lambda \cdot (C_{\text{measured}} - C_{\text{limit}}) \right]_+
\end{equation}

where $\alpha_\lambda$ is the dual learning rate and $[\cdot]_+ = \max(0, \cdot)$. This formulation ensures the agents converge to a policy that maximizes application throughput (FPS) while strictly adhering to the system's physical energy and latency budgets.


During Phase-C on-policy training, transitions are collected online at the granularity of complete perception chains (e.g., SAM or CLIP), and policy updates are performed after accumulating batches of 32 chain-level decisions, where one chain represents 6 tasks (3 for CLIP and 3 for SAM). The policy is optimized using Adam with a learning rate of $3\times10^{-4}$, a discount factor $\gamma=0.99$, a PPO clipping range of $0.2$, and two optimization epochs per update, with entropy regularization to promote exploration. This configuration balances learning stability with the real-time execution constraints of multi-robot systems.

\begin{figure}[b]
    \centering
    \begin{subfigure}[b]{0.155\textwidth}
        \centering
        \includegraphics[width=\linewidth]{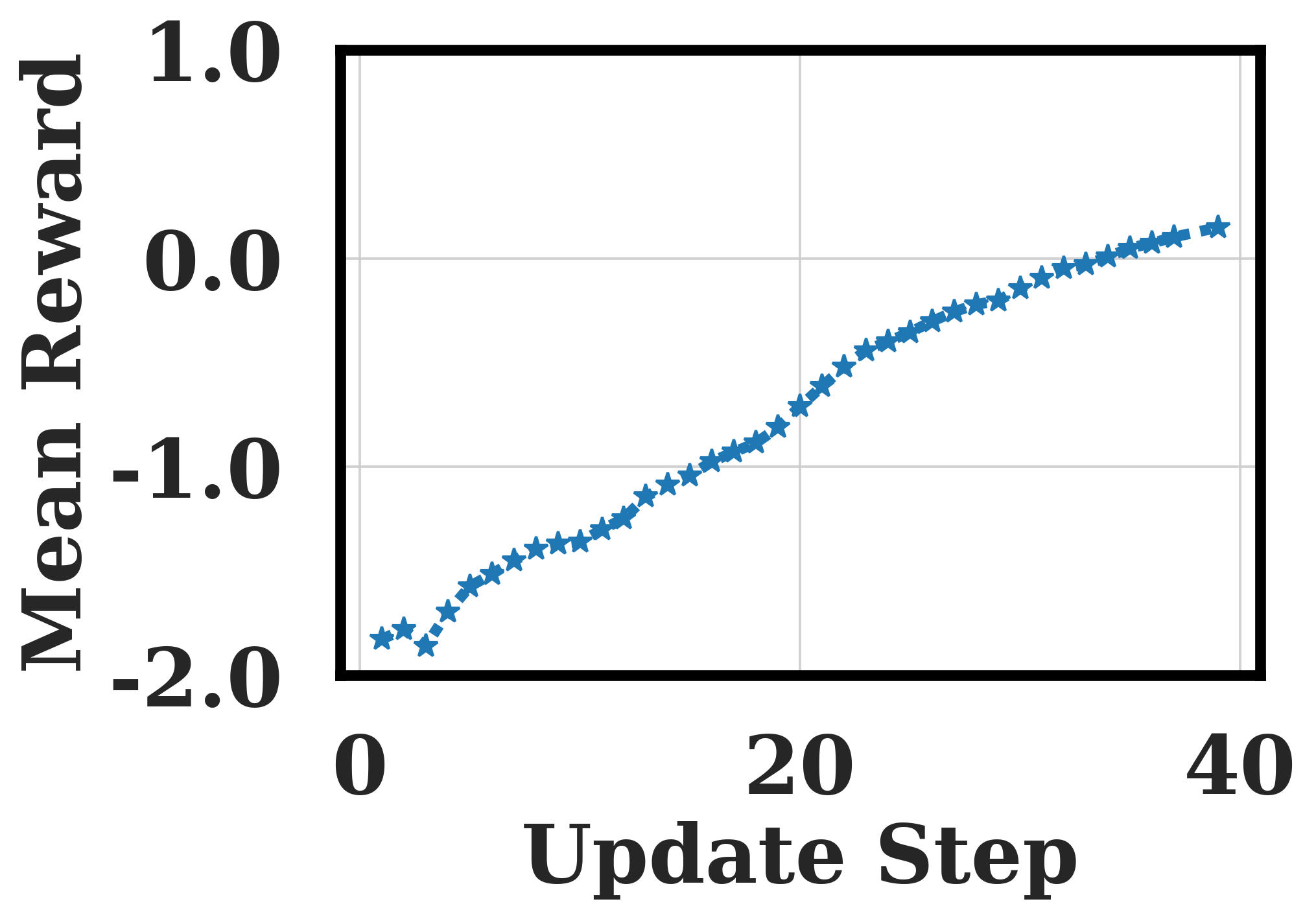}
        \caption{Spot}
        \label{fig:spot_reward}
    \end{subfigure}
    \hfill
    \begin{subfigure}[b]{0.155\textwidth}
        \centering
        \includegraphics[width=\linewidth]{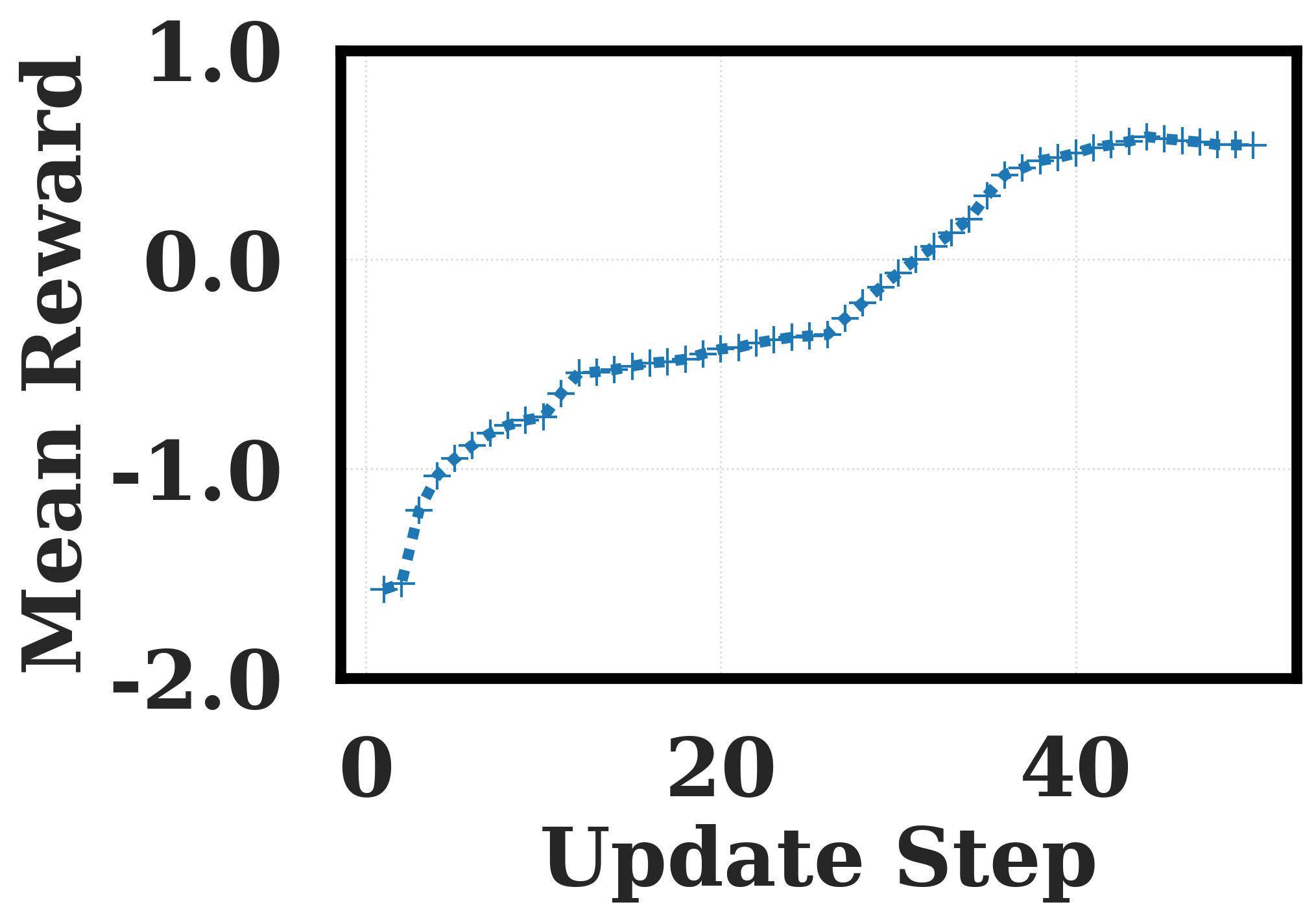}
        \caption{Jackal}
        \label{fig:jackal_reward}
    \end{subfigure}
    \hfill
    \begin{subfigure}[b]{0.155\textwidth}
        \centering
        \includegraphics[width=\linewidth]{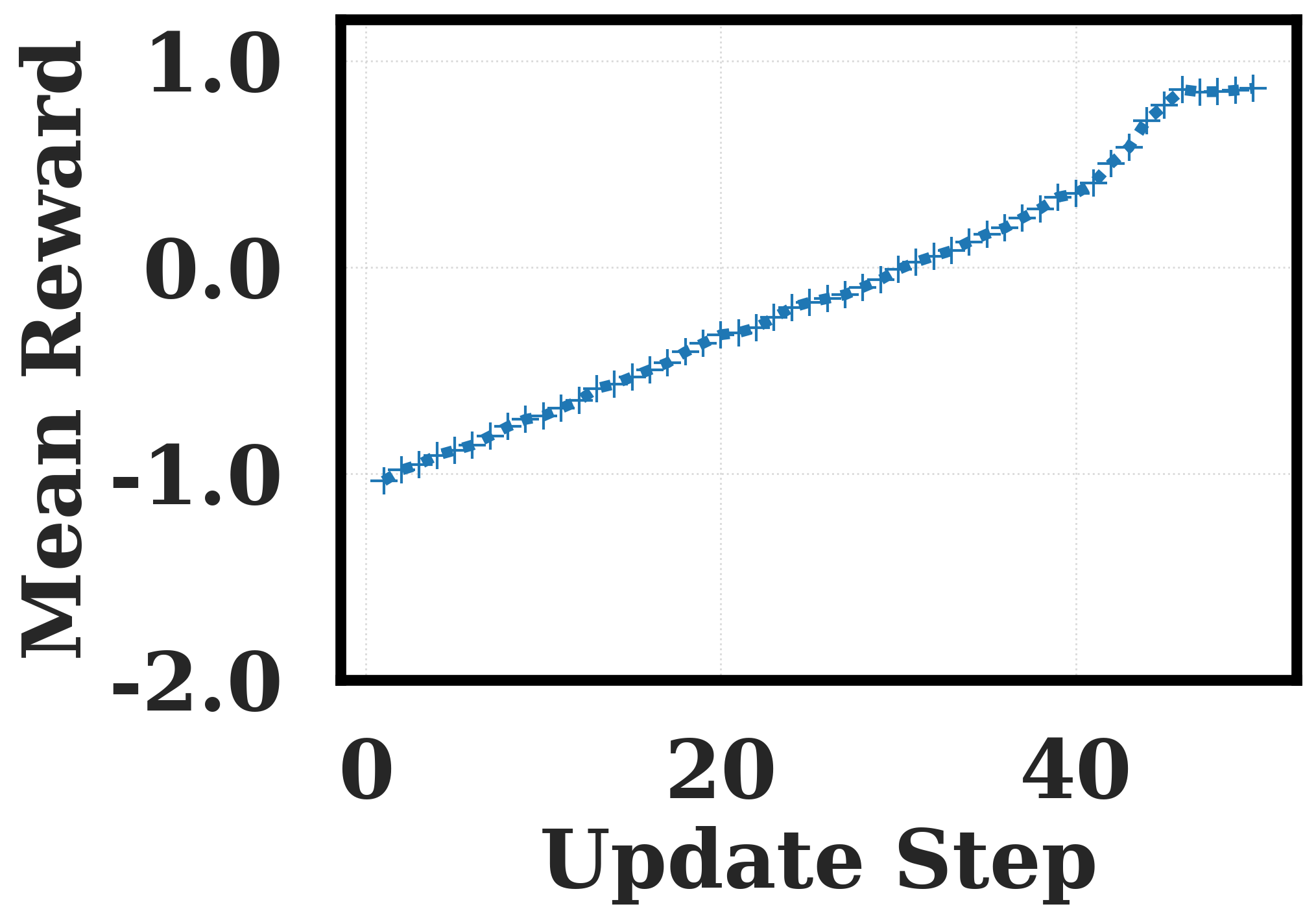}
        \caption{Husky}
        \label{fig:husky_reward}
    \end{subfigure}
    \caption{Mean reward progression across training updates for Spot, Jackal, and Husky under online reinforcement learning.}
    \label{fig:spot_jackal_husky_reward}
\end{figure}

Figure~\ref{fig:robot_loss_comparison} and figure~\ref{fig:spot_jackal_husky_reward} show the Phase-C training dynamics for Husky, Jackal, and Spot, reporting actor loss, critic loss, and mean reward versus update step. Across all robots, the actor loss decreases monotonically and converges to a stable regime, indicating consistent refinement of local versus auction-based execution decisions. The critic loss exhibits smooth convergence without divergence, demonstrating reliable value estimation despite delayed, chain-level rewards and stochastic execution latencies. The mean reward improves steadily from negative toward near-zero or positive values, reflecting increased deadline satisfaction and more effective offloading behavior. Husky converges more rapidly due to stronger onboard compute and more stable network conditions, while Jackal and Spot show slower yet stable convergence attributable to higher execution and communication variability. Overall, the results confirm stable and robust Phase-C on-policy learning across heterogeneous robotic platforms.

\section{Experimental Setting}
We evaluate the proposed framework on a heterogeneous multi-robot testbed under long-duration, real-world operation. The experimental design emphasizes comprehensive measurement of system behavior, capturing fine-grained task-level performance (FPS, latency, execution time), system-level resource utilization (CPU/GPU usage, memory, queue length), communication effects( round trip time(rtt), RSSI) , and battery dynamics (power, energy, and state-of-charge). This rich set of metrics enables a detailed analysis of how vision--language workloads interact with heterogeneous hardware and energy constraints.
\subsection{Communication Protocol} The framework utilizes ROS~2 over 5,GHz Wi-Fi for low-latency collaborative inference. Task modules exchange serialized representations via a publish/subscribe model supporting peer-to-peer and multicast execution, with result aggregation at resource-optimized nodes. Leveraging the DDS backend, the protocol enables dynamic discovery, allowing the RL scheduler to reconfigure assignments in real-time based on device and network states.

\subsection{Testbed}
 We evaluate the framework using a heterogeneous testbed comprising three robots and an external node, representing diverse mobility and compute profiles. The robotic platforms include: (1) a Clearpath Husky A300 AMP~\cite{husky} (high-capacity) equipped with a 13th-gen Intel i7 and NVIDIA RTX~3090; (2) a Clearpath Jackal~\cite{jackal} (mid-range) configured with an Intel i5 and GTX~1650~Ti; and (3) a Boston Dynamics Spot~\cite{spot} (legged) utilizing an onboard NVIDIA Orin for operation in unstructured terrain. To evaluate offloading scalability, we incorporate an Aorus 15P laptop (Intel i7-11800H, NVIDIA RTX 30-series 8,GB GPU) acting as a portable edge server. All systems run ROS~2.

\subsection{Data Collection For Offline Policy Training}
\begin{wrapfigure}{r}{0.65\linewidth}
    \centering
    \includegraphics[width=\linewidth]{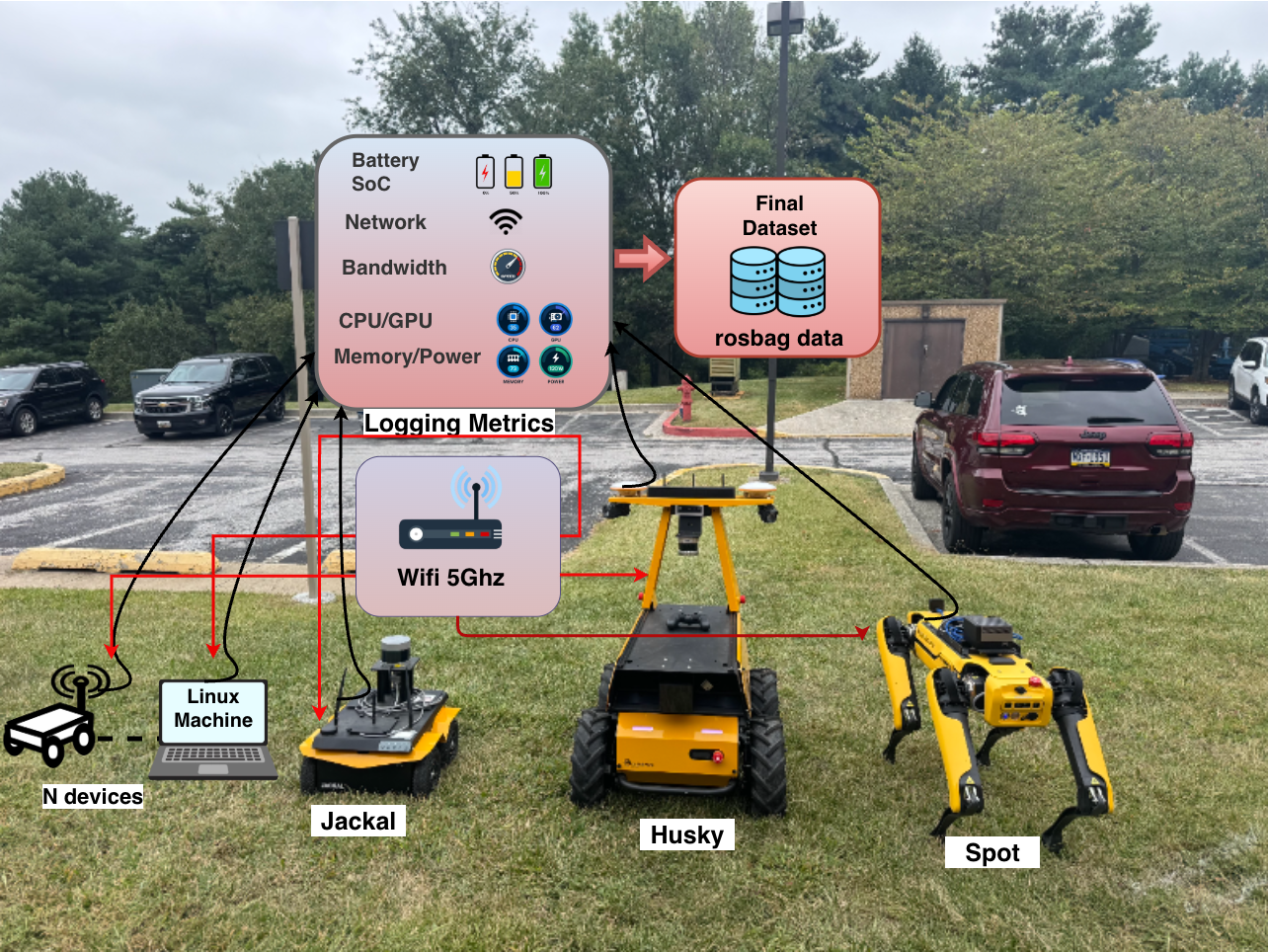}
    \caption{Multi-UGV Real-time Experimental Data Collection Pipeline}
    \label{fig:data_collection}
\end{wrapfigure}

We implement a systematic data collection framework ~\ref{fig:data_collection} that logs both system-level telemetry and task-level performance during continuous robot operation. Real-time battery data are recorded directly from ROS~2 topics specific to each platform. These topics provide direct measurements of voltage, current draw, power consumption, and state-of-charge (SoC), round trip time (rtt), RSSI, ensuring accurate characterization of energy usage under operational conditions. 
To evaluate the computational workload, we deploy multiple vision-language models (VLMs) across the heterogeneous testbed, specifically CLIP~\cite{radford2021clip} and Grounded-SAM~\cite{liu2023groundedsam}. Model tasks are executed on each UGV while logging per-task latency, total inference time, and estimated floating-point operations (FLOPs). Battery telemetry is captured concurrently, linking computational intensity with real-time energy demand.We observe offloading six tasks is expected to introduce approximately 140–180 ms of communication latency , depending on payload size and protocol overhead. In total, we collect more than 20 hours of data, encompassing model execution in both static and dynamic (in-motion) scenarios. This dataset captures a unique combination of perception-model workload metrics and mobility-induced energy consumption. By correlating FLOPs, latency, and inference throughput with real-time battery dynamics across heterogeneous robotic platforms, we create a benchmark that supports the evaluation of scheduling, workload distribution, and energy-aware autonomy in heterogeneous multi-robot systems.

\section{Results \& Discussion}
This section presents a comprehensive evaluation of the proposed framework under long-duration, real-world execution. We compare four scheduling strategies—\emph{Baseline}, \emph{Auction (Distributed Auction)}, \emph{GA (Genetic Algorithm)~\cite{9484436}}, and our Phase-C on-policy \emph{MAPPO} controller—across three heterogeneous robots (Husky, Jackal, and Spot) running concurrent CLIP and SAM perception pipelines. The Baseline represents fully local execution, where each robot independently processes all six CLIP and SAM task modules without sharing or offloading any computation to peer robots. The evaluation focuses on achieved FPS, end-to-end latency, task success rates, offloading behavior, and resource efficiency (CPU/GPU utilization, energy consumption, and battery SoC), providing insight into real-time performance.

\subsection{On-Policy Evaluation}

\begin{figure}[b]
    \centering

    \subfloat[Husky: FPS]{%
        \includegraphics[width=0.152\textwidth]{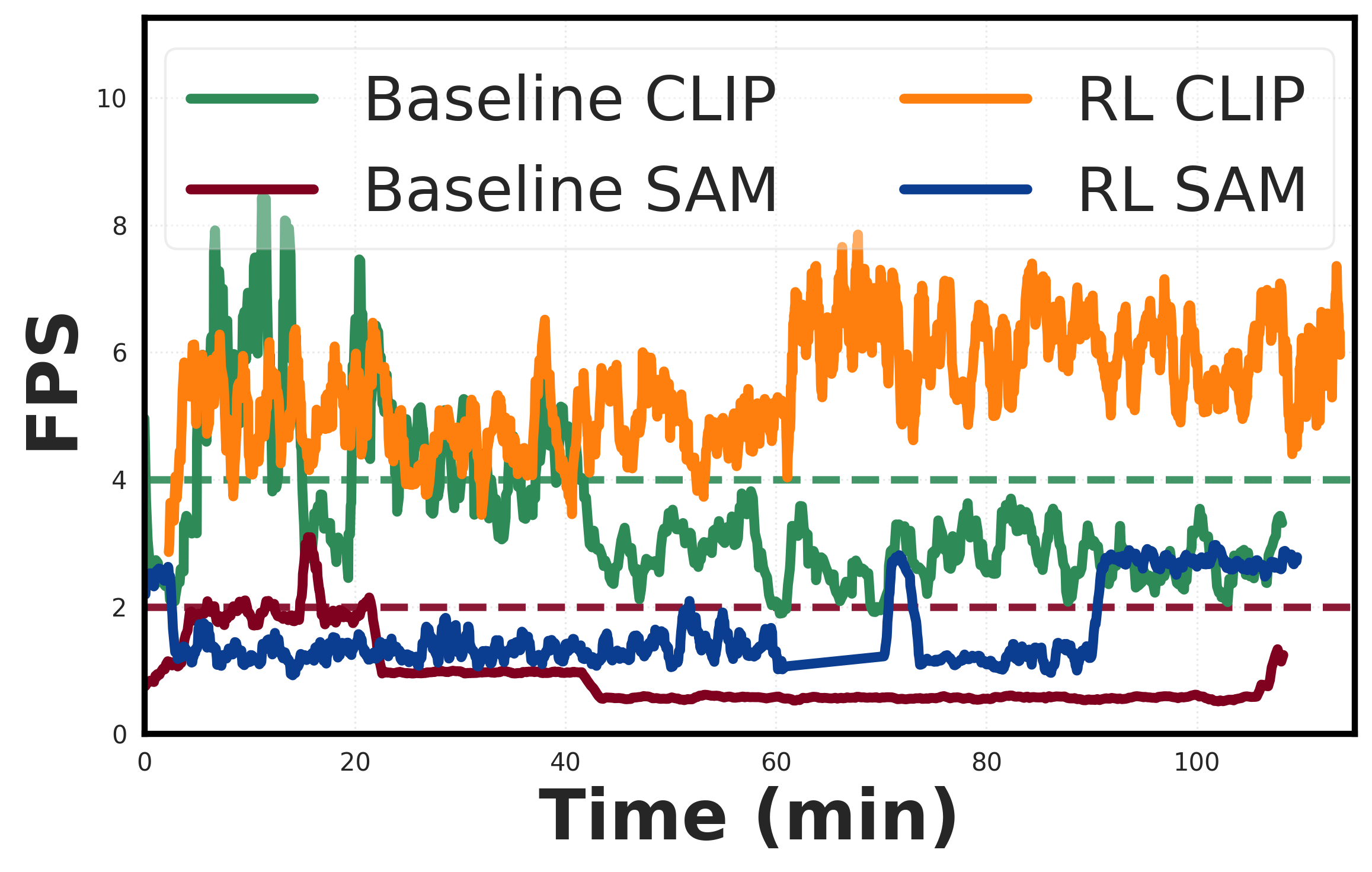}
    }\hfill
    \subfloat[Jackal: FPS]{%
        \includegraphics[width=0.148\textwidth]{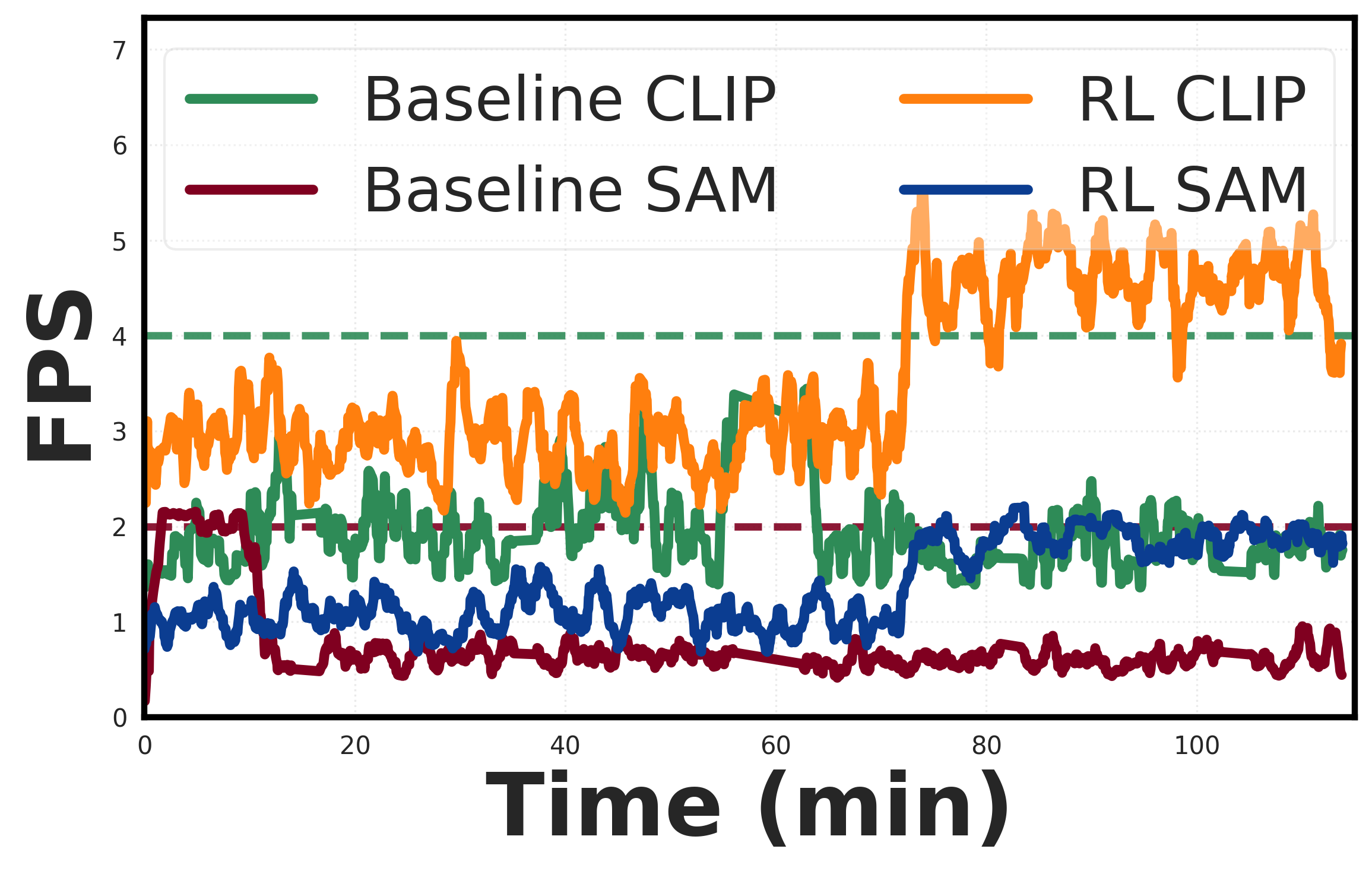}
    }\hfill
    \subfloat[Spot: FPS]{%
        \includegraphics[width=0.152\textwidth]{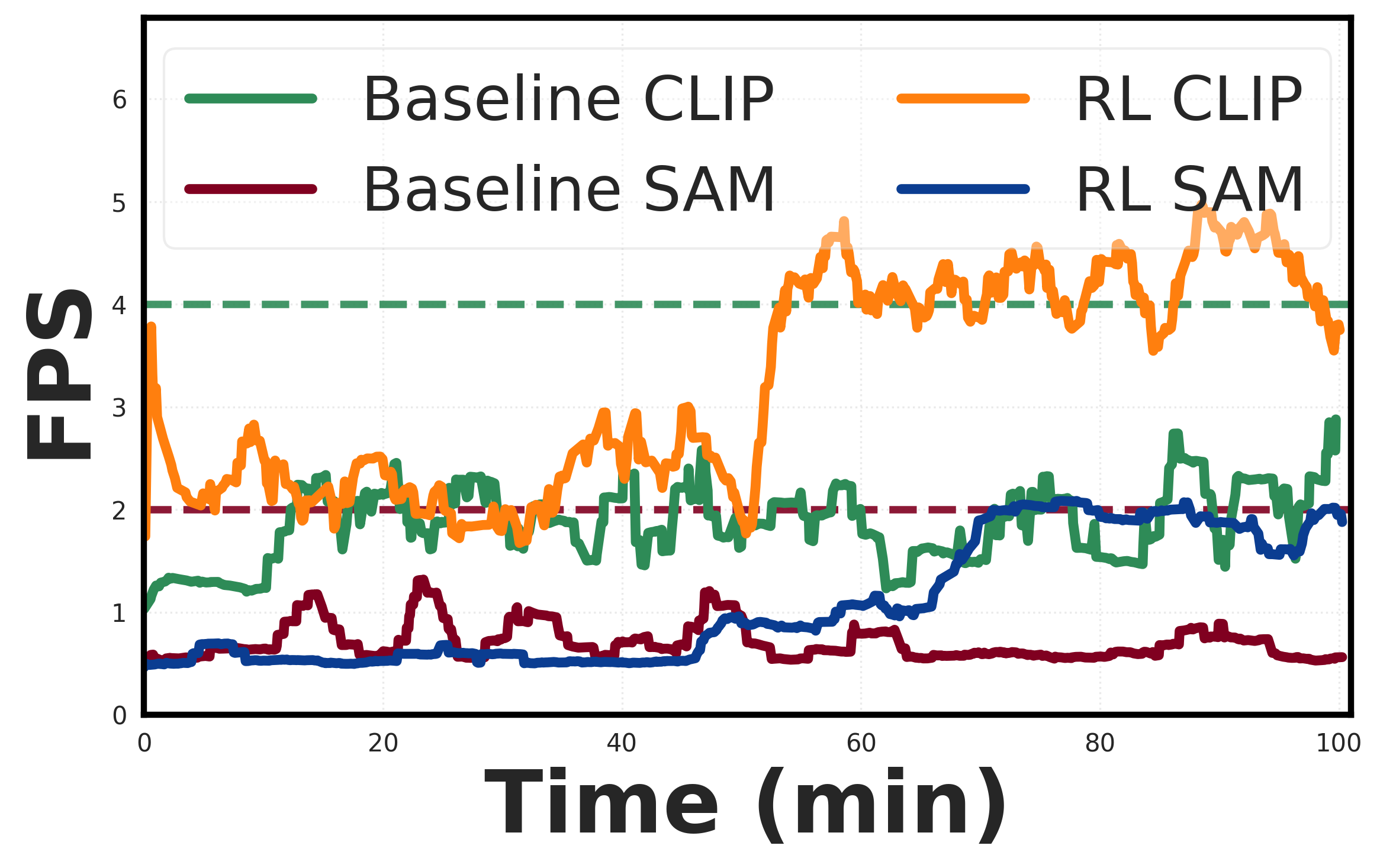}
    }\\[0.6em]

    \subfloat[Husky: Latency]{%
        \includegraphics[width=0.152\textwidth]{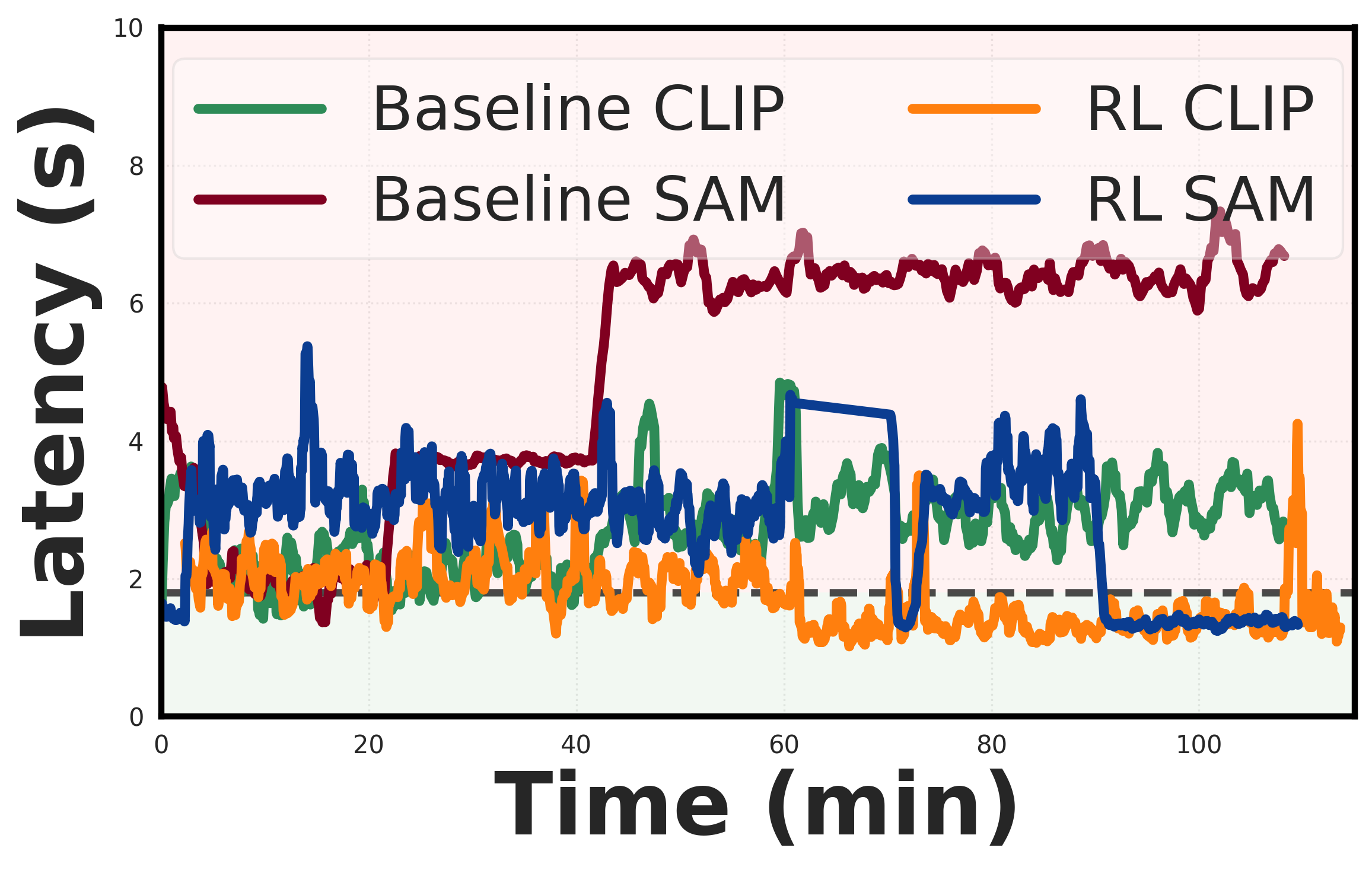}
    }\hfill
    \subfloat[Jackal: Latency]{%
        \includegraphics[width=0.151\textwidth]{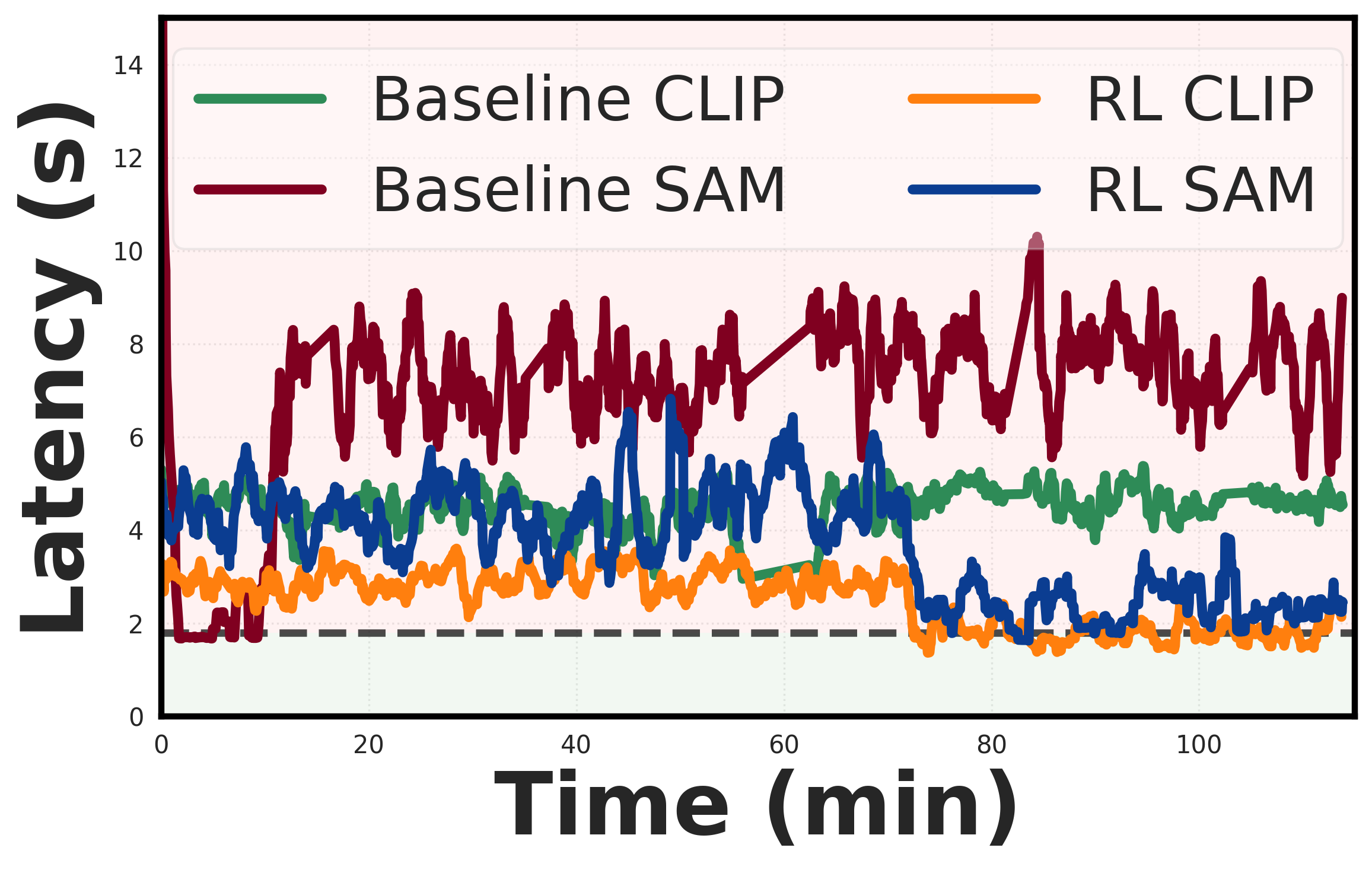}
    }\hfill
    \subfloat[Spot: Latency]{%
        \includegraphics[width=0.152\textwidth]{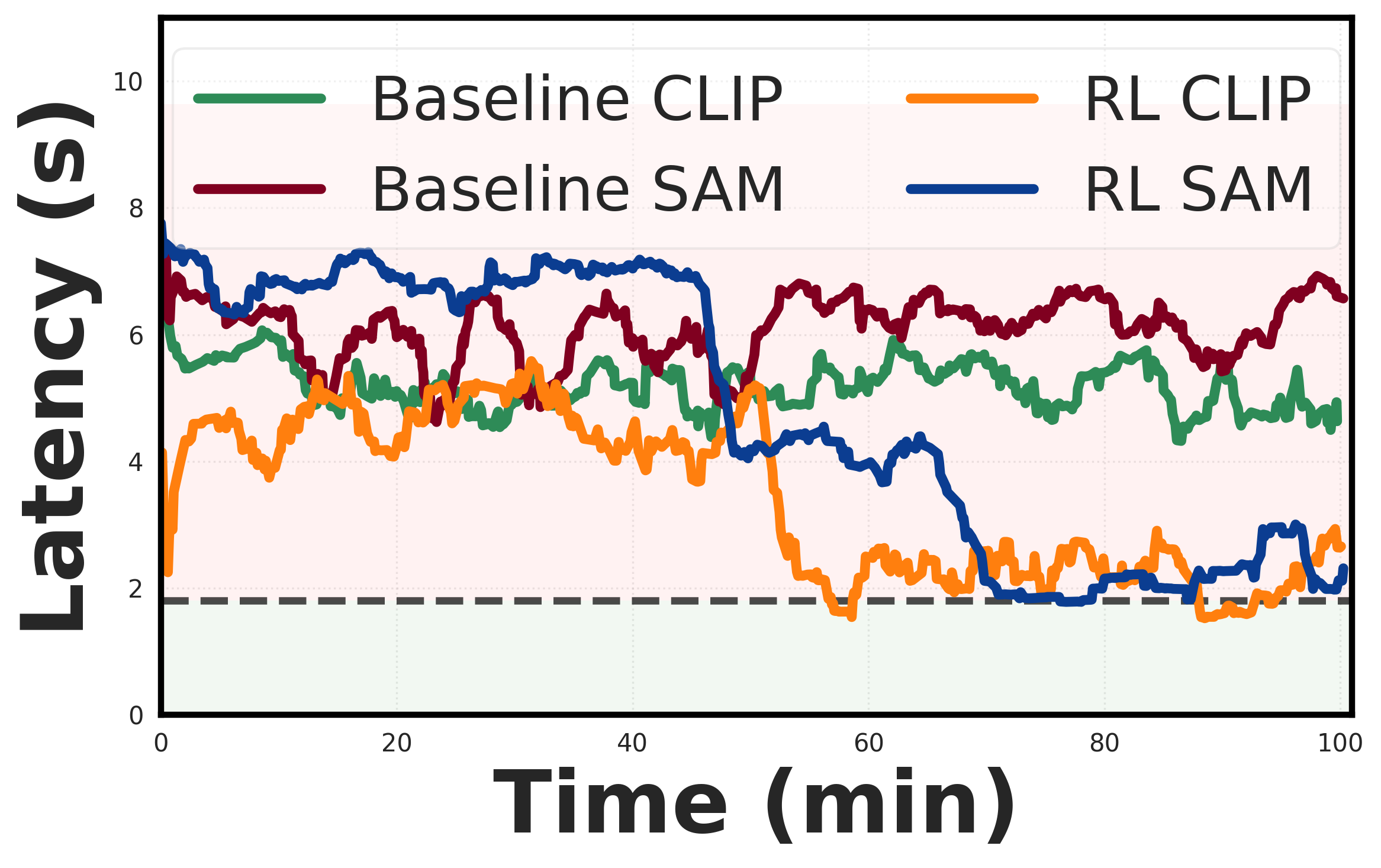}
    }

    \caption{Comparison of baseline and Phase-C on-policy MAPPO performance across Husky, Jackal, and Spot robots.
    For each platform, the achieved FPS (top) and achieved latency (bottom) are shown as temporal trajectories for both CLIP and SAM pipelines.
    Baseline execution (green and burgundy) is overlaid with the learned MAPPO policy (orange and blue), highlighting the effect of learning-based scheduling.
    Across all robots, the MAPPO controller reduces performance jitter, keeps achieved FPS closer to the target, and stabilizes latency relative to the baseline,
    while adapting to heterogeneous compute capabilities and long-duration operation.}
    \label{fig:Baseline_vs_RL}
\end{figure}

Figure~\ref{fig:Baseline_vs_RL} reports on-policy evaluation on three heterogeneous robots (Husky, Jackal, Spot), comparing the baseline scheduler against our Phase-C MAPPO controller for both CLIP and SAM. The baseline frequently operates outside the desired QoS region: CLIP often falls below the 4~FPS goal and exhibits sustained latency above the 1.8~s bound, while the heavier SAM chain is consistently more constrained, with low throughput (typically well below 2~FPS) and large latency excursions (often several seconds, and in some runs exceeding $\sim$6--9~s). This confirms that without learning-based coordination, the system spends most of its time missing at least one constraint, particularly for SAM.

In contrast, \names shifts both pipelines toward the target regime and noticeably damps long-horizon variability. On Husky, the MAPPO policy maintains higher and more stable CLIP throughput (sustained above the 4~FPS line for extended intervals) while reducing CLIP latency into a tighter band close to the target bound; SAM similarly improves, avoiding prolonged low-FPS operation and exhibiting lower, more bounded latency compared to the baseline. On Jackal, MAPPO yields a clear step change mid-run: CLIP throughput increases toward the 4~FPS target and both CLIP and SAM latencies drop into a lower, more stable regime, while SAM throughput rises toward its 2~FPS goal. Spot shows the strongest contrast—baseline trajectories remain in a low-throughput/high-latency regime, whereas MAPPO raises CLIP throughput toward (and often above) 4~FPS and progressively improves SAM throughput, while driving both chains’ latencies down toward the 1.8~s bound. Because Phase-C uses on-policy MAPPO, these improvements emerge from continual updates using fresh rollouts collected during execution, allowing the policy to adapt online to non-stationary contention, platform heterogeneity, and long-duration operation.

\begin{figure}[b]
    \centering
    \begin{minipage}{0.235\textwidth}
        \centering
        \includegraphics[width=\linewidth]{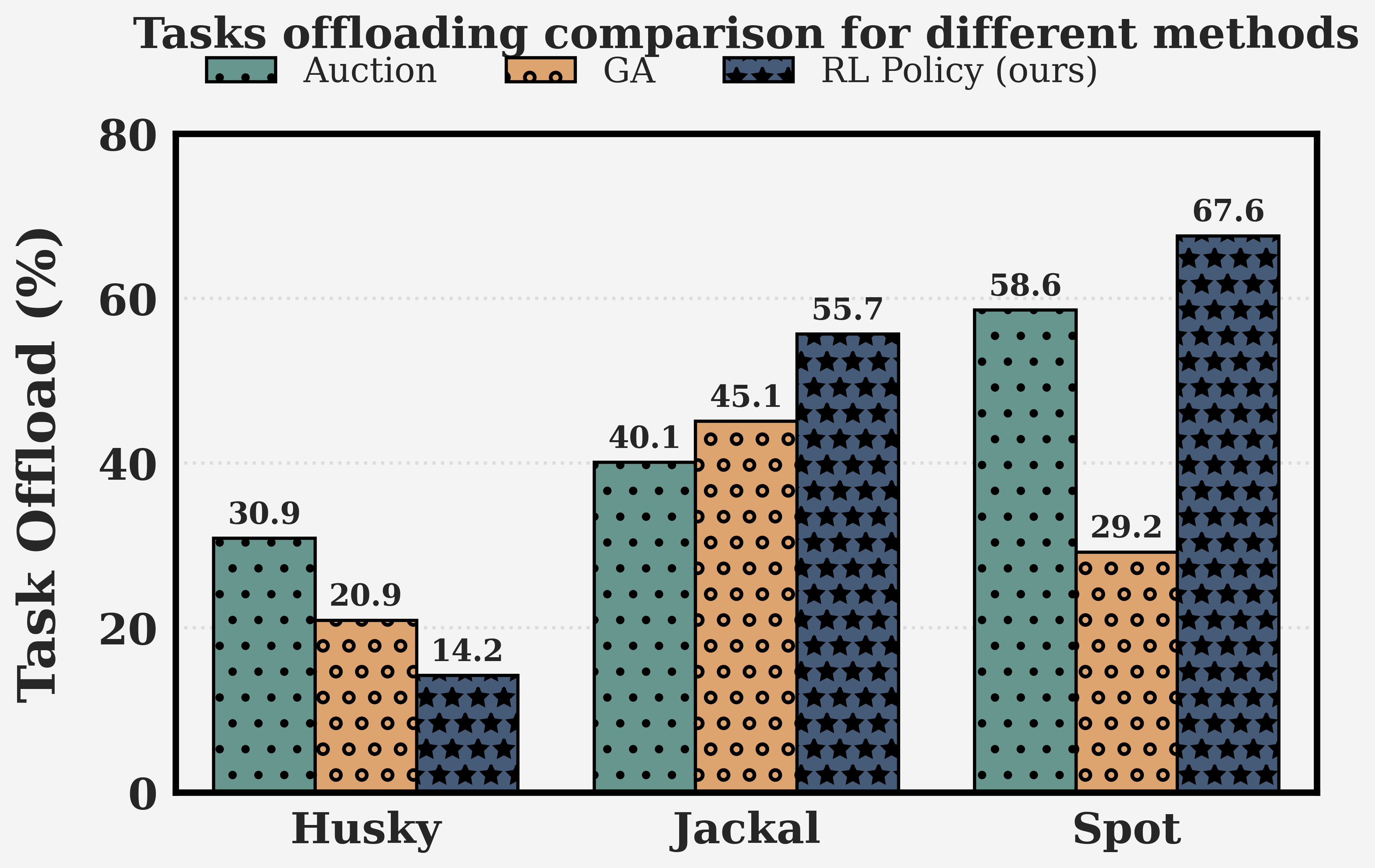}
        \caption{Comparison of task offloading for Auction, GA, and our RL policy across different robots.}
        \label{fig:offload_comparison}
    \end{minipage}
    \hfill
    \begin{minipage}{0.235\textwidth}
        \centering
        \includegraphics[width=\linewidth]{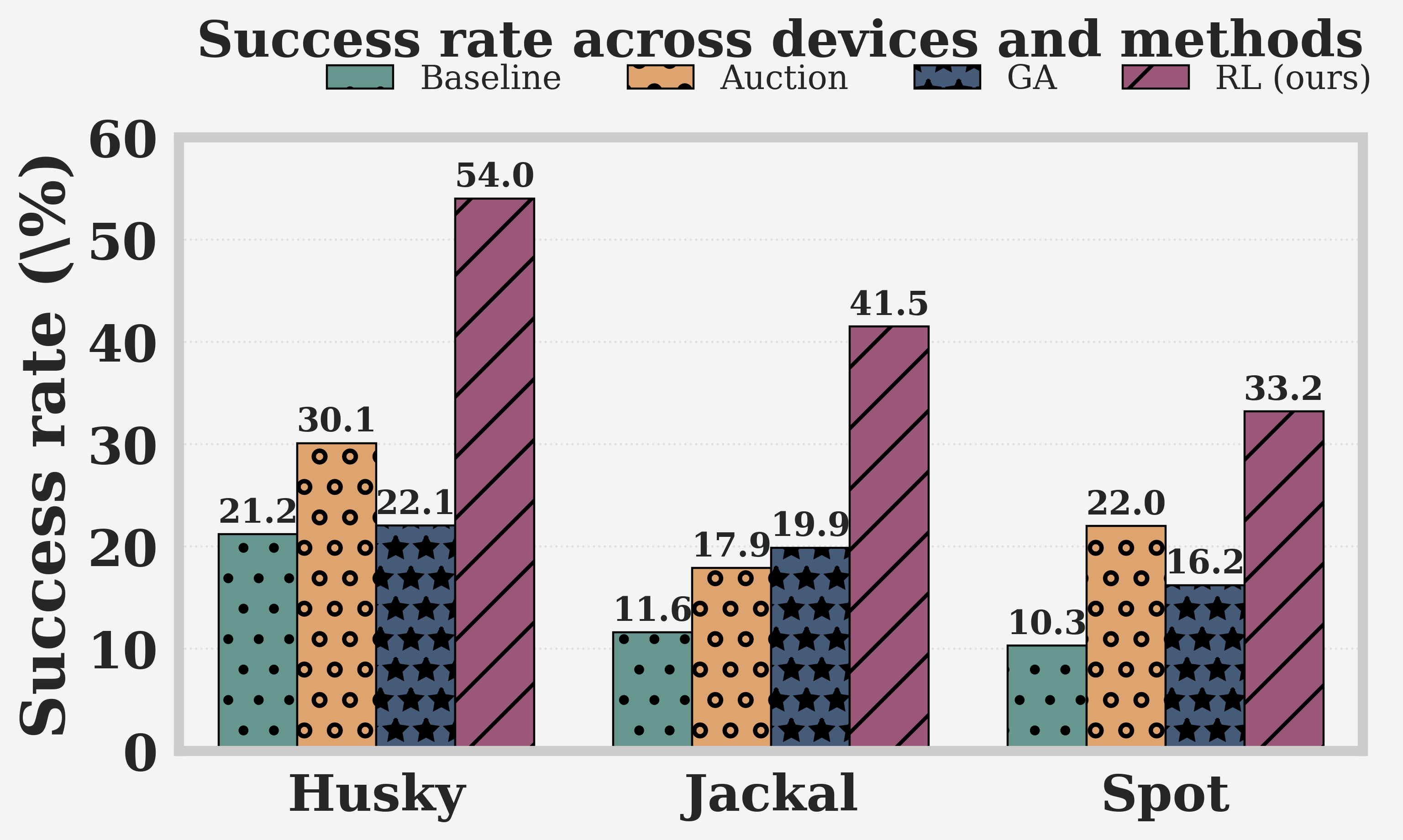}
        \caption{Success rate of meeting goal FPS and latency budget across robots and scheduling strategies. The RL policy consistently outperforms the baseline, auction-based, and GA schedulers on all three platforms.}
        \label{fig:success_rate}
    \end{minipage}
\end{figure}

Figure~\ref{fig:success_rate} summarizes the percentage of sam and clip chains that simultaneously satisfy the goal FPS and end-to-end latency budget for all devices and scheduling strategies. The Baseline runs all tasks locally on each robot without sharing; the Auction method assigns stages to the lowest-cost bidder; GA denotes a genetic-algorithm heuristic over auction bids; and RL is
our learned policy. 

Across all three robots, the RL scheduler is consistently the most reliable. On Husky, RL attains a success rate of \textbf{54.0\%}, compared to \textbf{21.21\%} (Baseline), \textbf{30.1\%} (Auction), and \textbf{22.1\%} (GA), i.e., about $2.5\times$ the Baseline and roughly $1.8\times$  the performance of best non-learning method ( auction). On Jackal, RL reaches \textbf{41.5\%} versus \textbf{11.6\%}, \textbf{17.9\%}, and \textbf{19.9\%}, yielding more than $3.5\times$ improvement over the Baseline and over $2\times$ the Auction scheduler. On Spot, RL achieves \textbf{33.2\%}, compared to \textbf{10.3\%}, \textbf{22.0\%}, and \textbf{16.2\%}, corresponding to $\sim 3.2\times$ the
Baseline and $\sim 50\%$ gain over Auction.

Overall, collaborative offloading (Auction, GA, RL) clearly outperforms the fully local Baseline, confirming the benefit of sharing workloads across robots. However, the relatively modest and sometimes inconsistent gains of GA over the cost-based Auction, compared with the consistently large margins of RL, highlight that learning-based scheduling is crucial for robust QoS guarantees in heterogeneous multi-robot systems.

Figure~\ref{fig:offload_comparison} compares task-offloading behavior under three methods: Auction, GA, and our RL policy. The Auction baseline exhibits moderate offloading rates but reacts only to instantaneous cost estimates, leading to inconsistent offload patterns across heterogeneous robots. The GA-based scheduler, which optimizes over candidate schedules without online adaptation, tends to be conservative, resulting in reduced offloading for Husky and Spot and only slight improvement for Jackal. Both non-learning methods lack awareness of temporal trends, workload bursts, and robot-specific compute bottlenecks.


In contrast, our RL policy demonstrates clear adaptive behavior shaped by resource conditions. The RL agent minimizes unnecessary offloading on Husky (14.2\%), whose GPU can sustain the vision pipeline locally, thereby reducing communication overhead. For Jackal and Spot---devices that more frequently experience CPU/GPU saturation---the RL policy increases offloading to 55.7\% and 67.6\%, respectively. This selective behavior indicates that the agent learns \emph{when} offloading improves latency and \emph{when} it unnecessarily burdens the network. Overall, compared to the non-learning baselines, the RL policy achieves a more efficient and context-aware workload distribution, aligning decisions with each robot's hardware profile, live resource usage, and latency budgets, and thus enabling more stable collaboration and better utilization of shared compute resources across the fleet.

Table~\ref{tab:resource_usage_70min} compares CPU/GPU utilization, energy consumption, and state-of-charge (SoC) drop across three robots under Baseline, GA, Auction, and RL scheduling over a fixed 70-minute window.
On Husky, GA lowers CPU utilization (15.94\%~$\rightarrow$~10.94\%) but increases SoC drop (25\%~$\rightarrow$~27\%), while the Auction method raises both compute utilization and energy (166.22~Wh) without improving SoC (26.3\%). In contrast, the RL policy drives higher GPU usage (46.20\%) yet reduces SoC drop to 20.5\%. The smaller SoC loss indicates more effective use of available battery capacity, likely due to fewer idle stalls and more consistent workload execution. indicating more effective battery utilization despite higher energy draw (175.7~Wh).

\begin{table}[b]
\caption{CPU/GPU utilization, energy consumption, and SoC drop over a fixed evaluation window.}
\label{tab:resource_usage_70min}
\centering
\setlength{\tabcolsep}{3pt}
\renewcommand{\arraystretch}{1.0}
\resizebox{\columnwidth}{!}{%
\begin{tabular}{llcccc}
\hline
\textbf{Robot} & \textbf{Method} & \textbf{CPU (\%)} & \textbf{GPU (\%)} & \textbf{Energy (Wh)} & \textbf{SoC drop (\%)}\\
\hline
Husky  & Baseline & 15.94 & 22.84 & 144.30 & 25.00\\
       & GA       & 10.94 & 33.73 & 147.80 & 27.00\\
       & Auction  & 18.36 & 28.60 & 166.22 & 26.30\\
       & RL Policy (ours) & 23.80 & 46.20 & 175.70 & 20.50\\
\hline
Jackal & Baseline & 23.54 & 50.56 & 34.89  & 27.43\\
       & GA       & 27.67 & 55.20 & 27.28  & 25.60\\
       & Auction  & 28.65 & 48.20 & 30.55  & 25.10\\
       & RL Policy (ours) & 25.38 & 62.62 & 31.85 & 21.74\\
\hline
Spot   & Baseline & 33.00 & 45.00 & 216.00 & 67.00\\
       & GA       & 46.80 & 47.67 & 232.10 & 64.80\\
       & Auction  & 34.60 & 43.82 & 210.60 & 68.70\\
       & RL Policy (ours) & 35.00 & 58.00 & 198.00 & 62.00\\
\hline
\end{tabular}}
\end{table}
On Jackal, GA reduces energy (34.89~Wh~$\rightarrow$~27.28~Wh) and SoC drop (27.43\%~$\rightarrow$~25.60\%), whereas Auction provides limited additional benefit (25.1\% SoC). The RL policy further lowers SoC drop to 21.74\% while keeping energy below the baseline.
For Spot, GA slightly improves SoC drop at the cost of higher energy, and Auction lowers energy (210.6~Wh) but still incurs a high SoC drop (68.7\%). The RL policy achieves the best trade-off, reducing both energy (198~Wh) and SoC drop (62\%).
Overall, while the GA and Auction strategies can outperform the baseline in isolated cases, their benefits are inconsistent across
platforms. The RL policy consistently achieves the lowest SoC drop across all robots and often reduces total energy consumption and better resource utilization, highlighting the advantage of learning-based scheduling in jointly
optimizing compute utilization and battery health in heterogeneous multi-robot systems.

\subsection{Scalability}
To evaluate the scalability of the proposed RL-based task scheduler, we study its behavior under increasing system scale, resource heterogeneity, and workload intensity. When a new device is added, it does not require retraining from scratch; instead, it is initialized by masking the RL policy of an existing robotic agent with similar computational capability, allowing it to participate immediately in the auction process. In our experiments, the added Linux executor has CPU/GPU characteristics comparable to those of Husky and therefore adopts Husky’s learned policy. This policy masking allows the system to scale without architectural changes or additional training overhead. The following experiments assess learning convergence, runtime performance, and robustness as the number of devices and workload complexity vary.

\begin{figure}[b]
    \centering
    \begin{subfigure}[b]{0.155\textwidth}
        \centering
        \includegraphics[width=\linewidth]{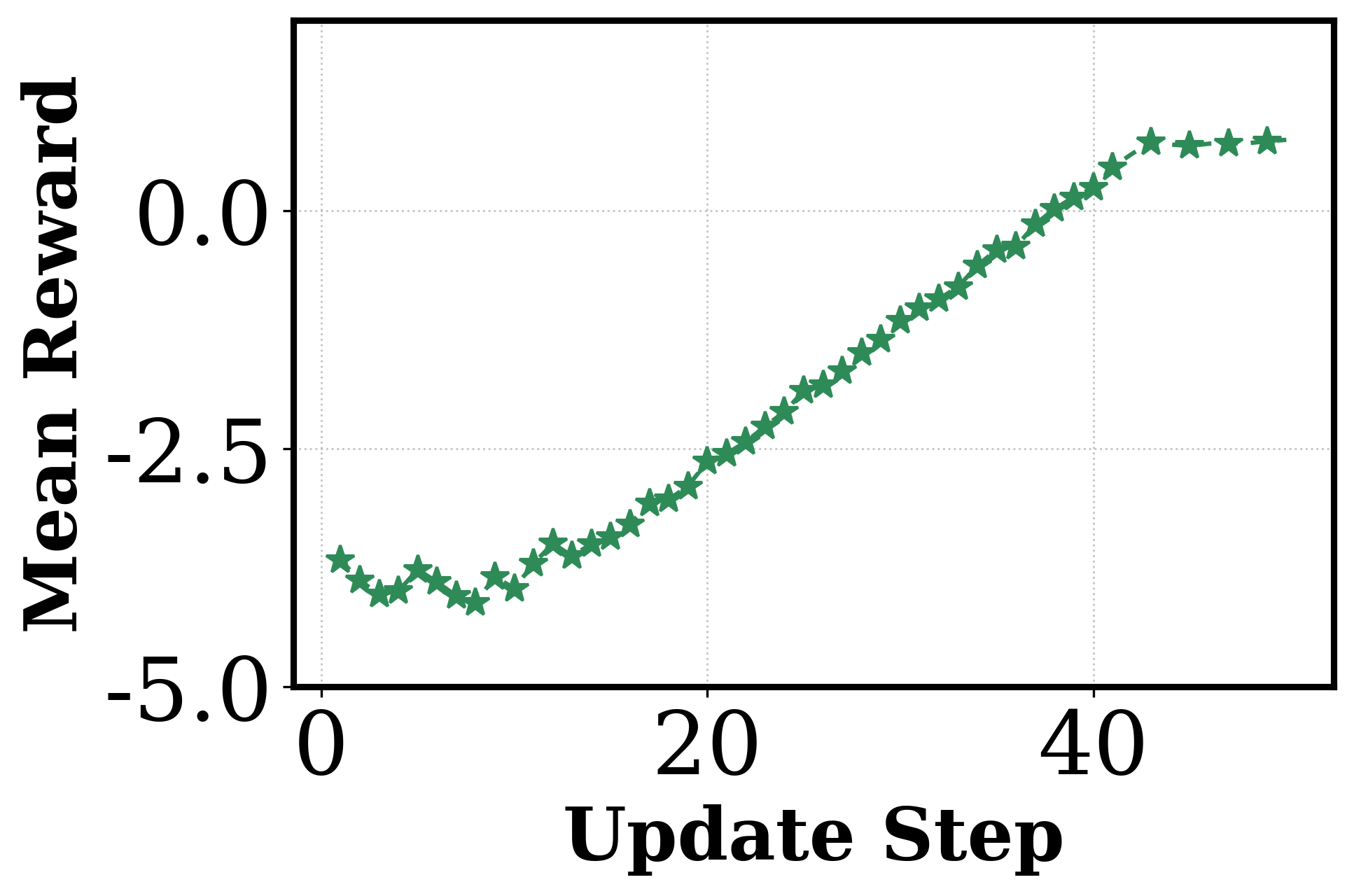}
        \caption{Spot}
        \label{fig:spot_reward_fourth}
    \end{subfigure}
    \hfill
    \begin{subfigure}[b]{0.155\textwidth}
        \centering
        \includegraphics[width=\linewidth]{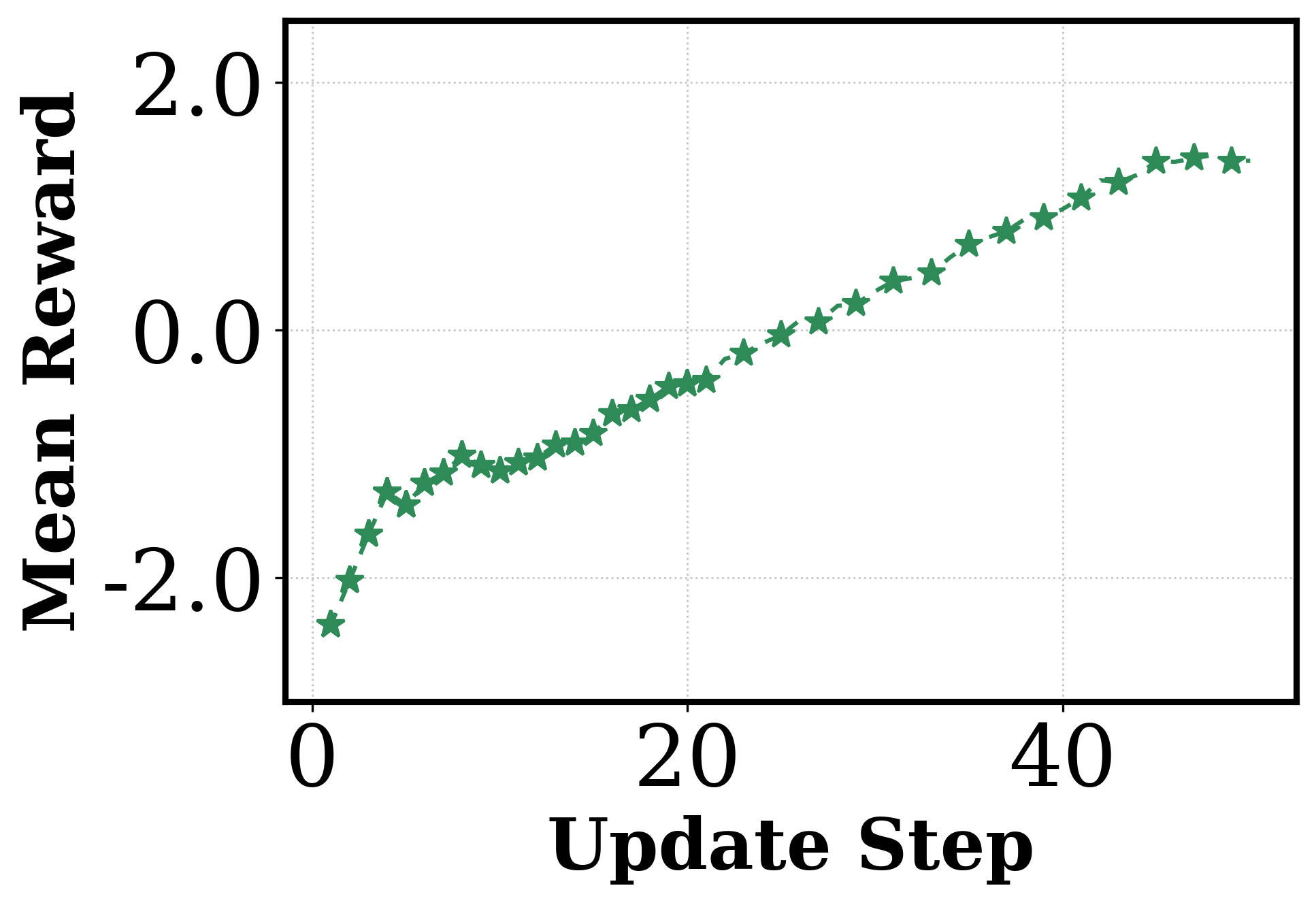}
        \caption{Jackal}
        \label{fig:jackal_reward_fourth}
    \end{subfigure}
    \hfill
    \begin{subfigure}[b]{0.155\textwidth}
        \centering
        \includegraphics[width=\linewidth]{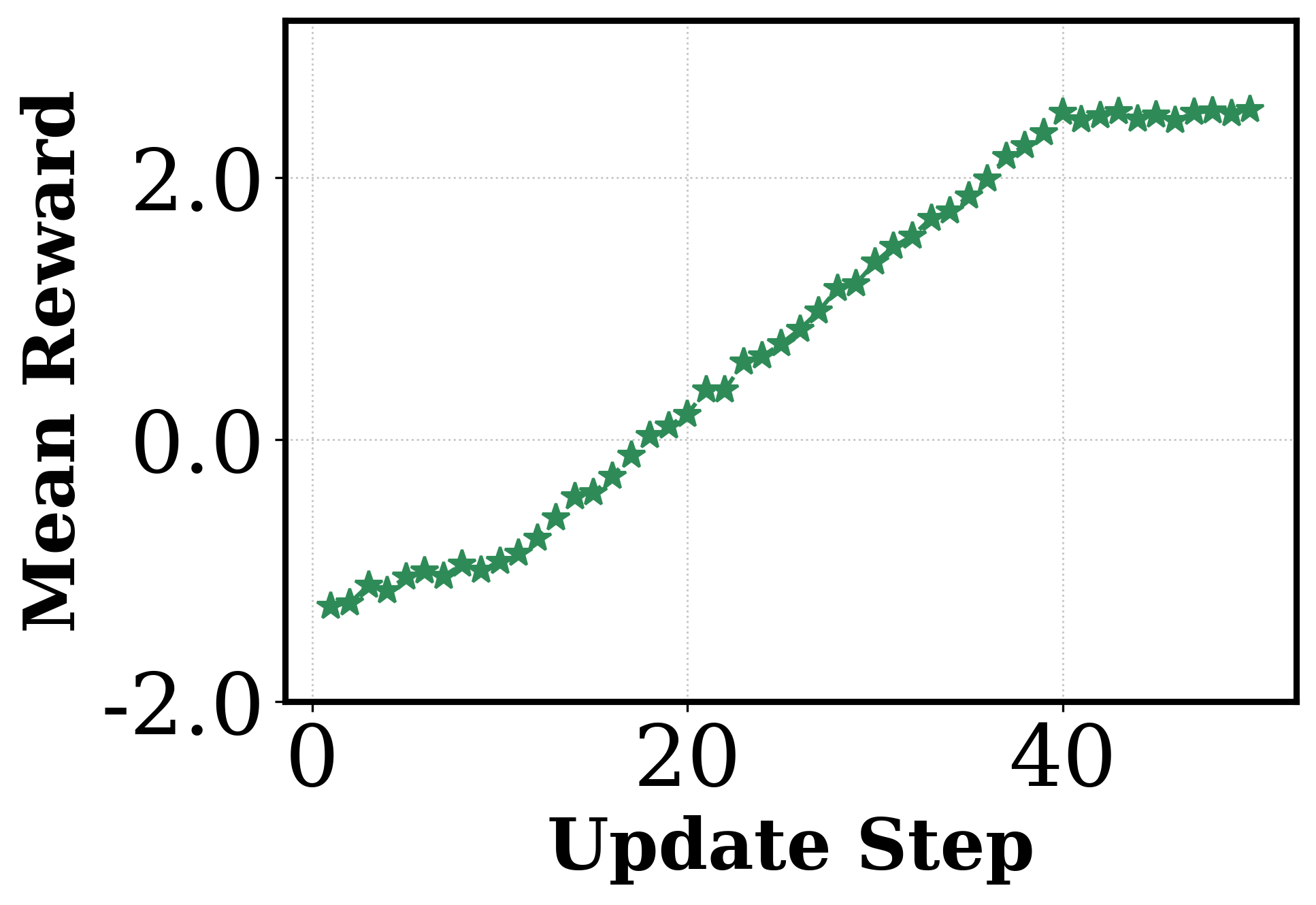}
        \caption{Husky}
        \label{fig:husky_reward_fourth}
    \end{subfigure}
    \caption{Four-device setup: on-policy RL convergence. Mean reward over 50 updates for Spot, Jackal, and Husky shows stable learning and saturation after ~40–45 updates under heterogeneous system conditions.}
    \label{fig:spot_jackal_husky_reward_fourth}
\end{figure}

\begin{figure}[b]
    \centering

    \subfloat[Husky: FPS]{%
        \includegraphics[width=0.152\textwidth]{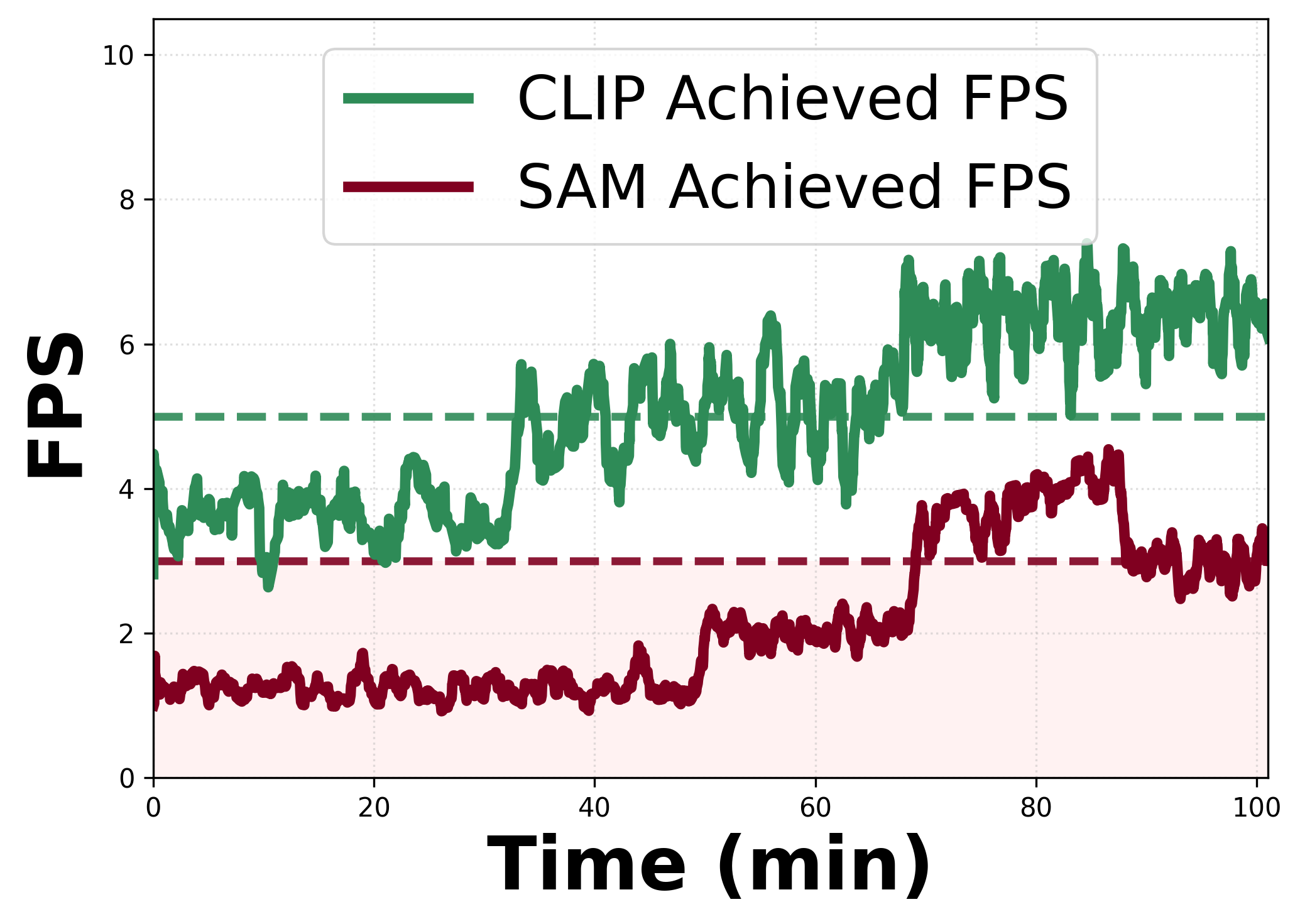}
    }\hfill
    \subfloat[Jackal: FPS]{%
        \includegraphics[width=0.148\textwidth]{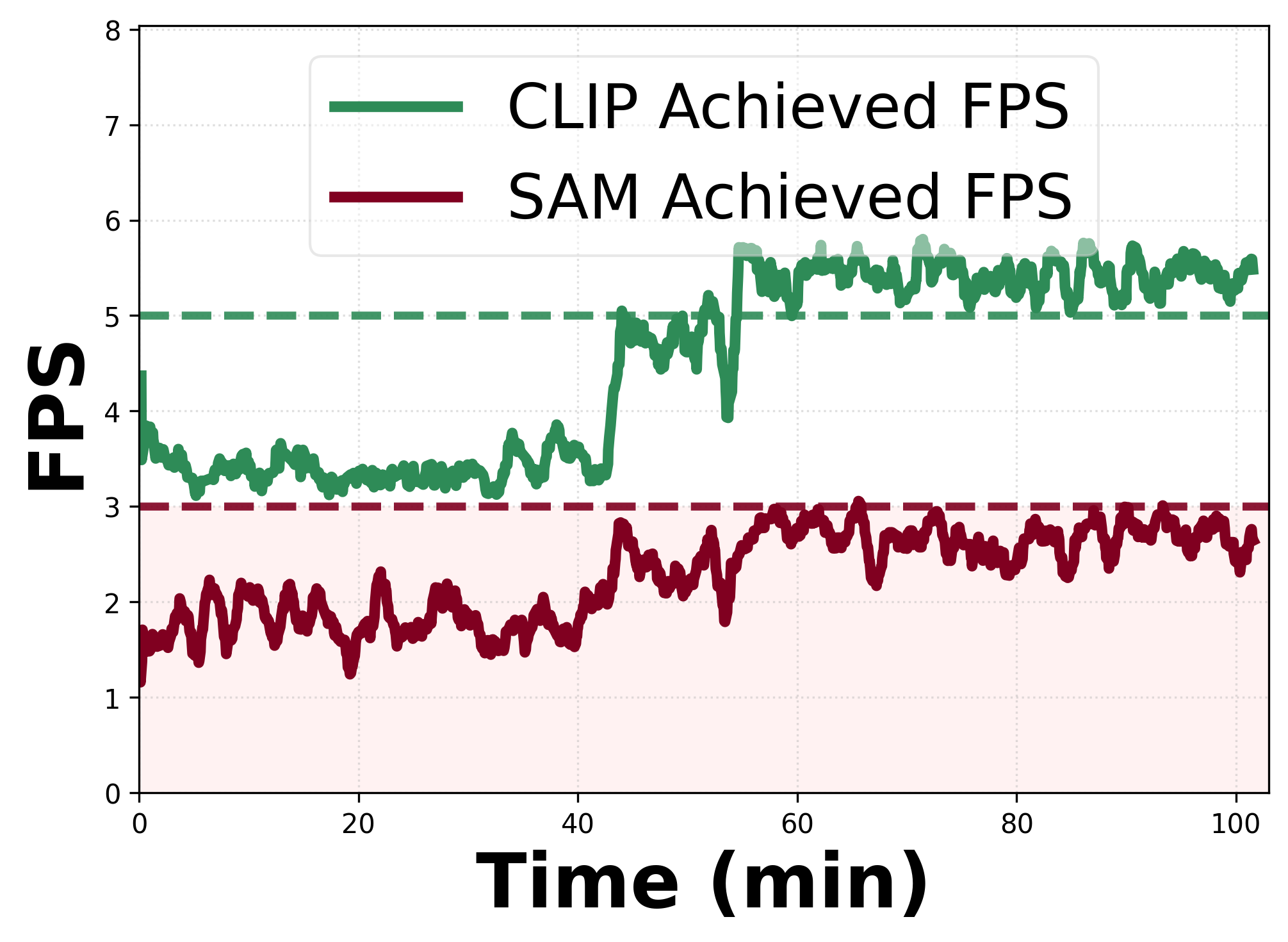}
    }\hfill
    \subfloat[Spot: FPS]{%
        \includegraphics[width=0.152\textwidth]{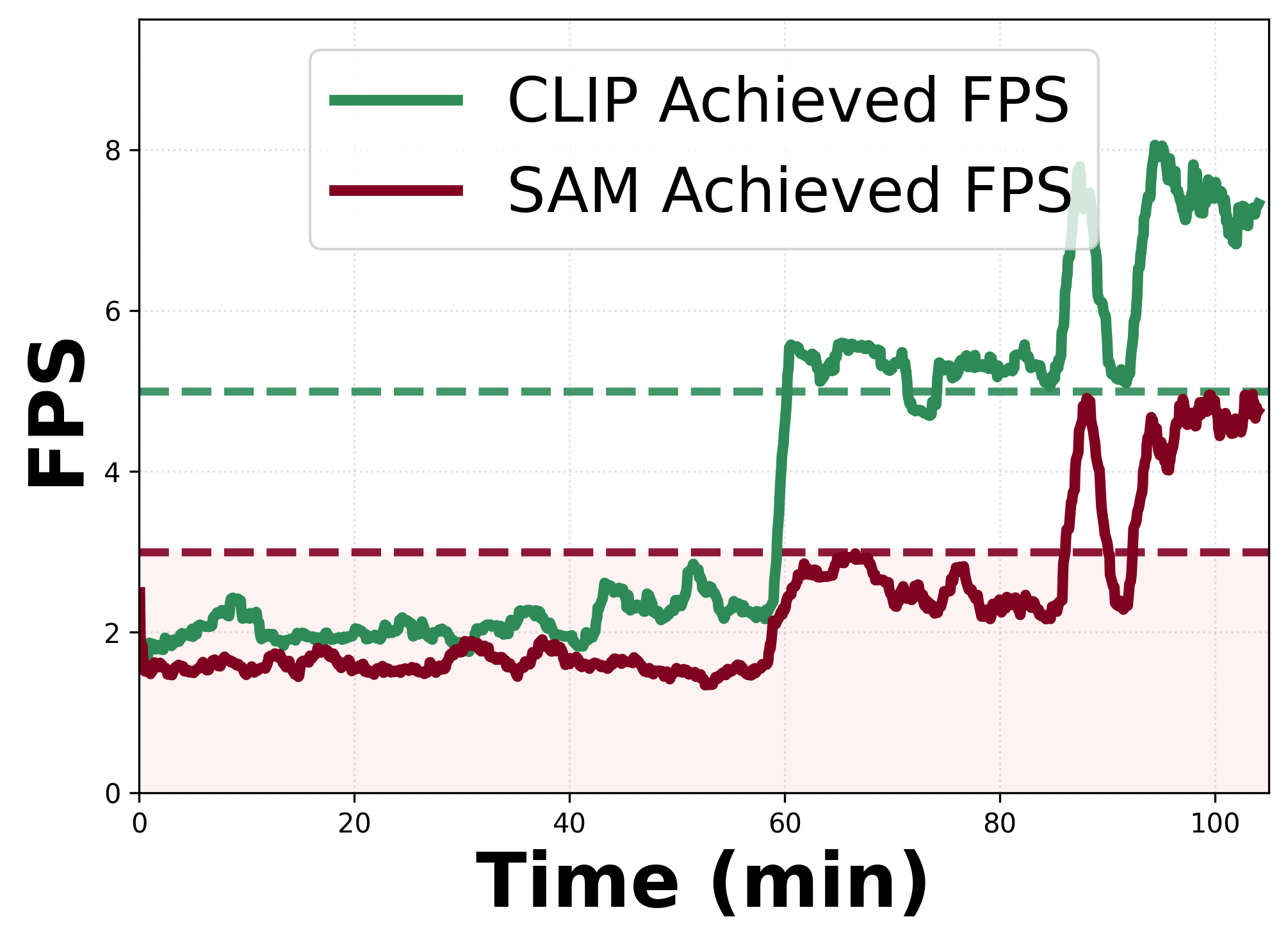}
    }\\[0.6em]

    \subfloat[Husky: Latency]{%
        \includegraphics[width=0.152\textwidth]{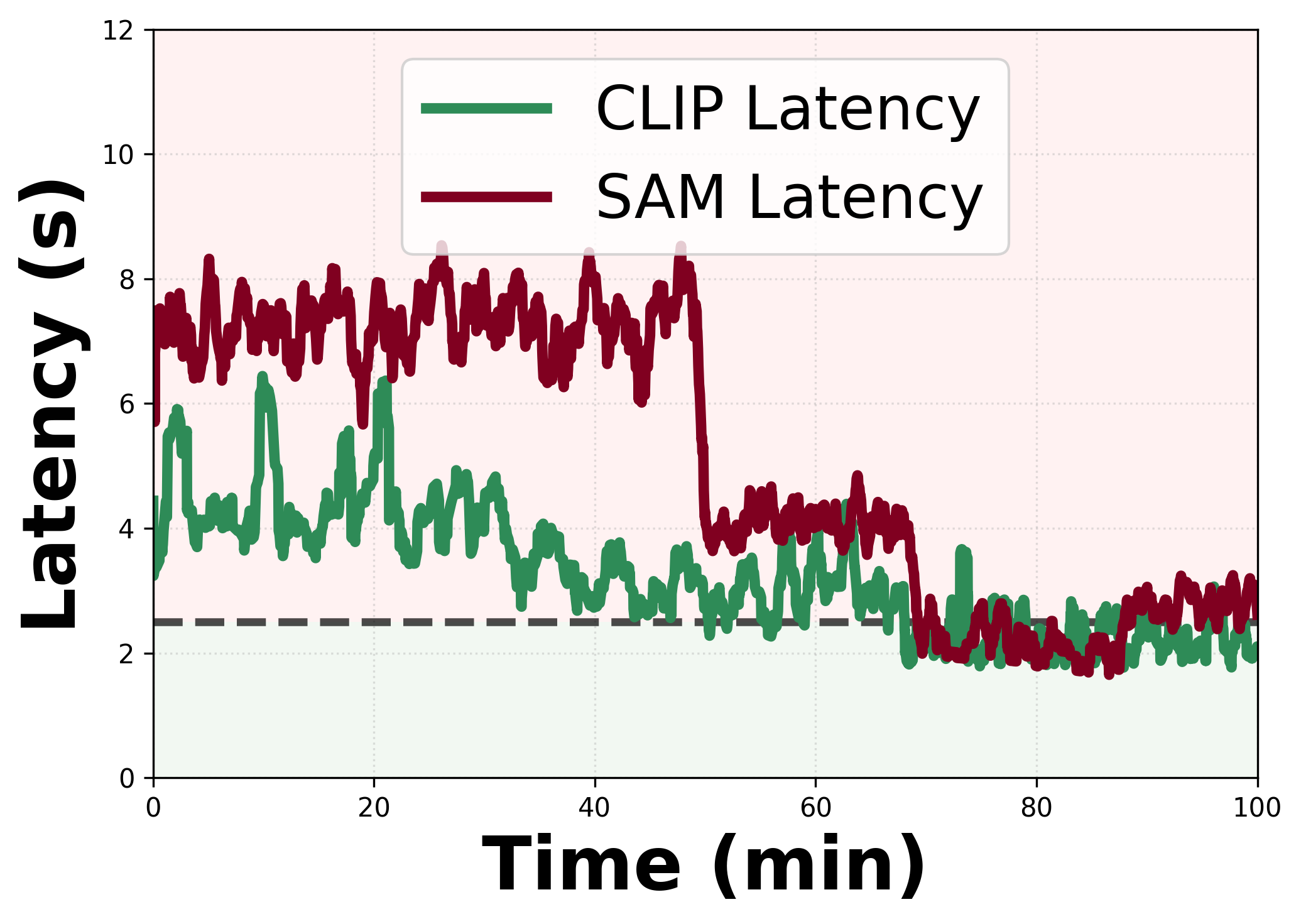}
    }\hfill
    \subfloat[Jackal: Latency]{%
        \includegraphics[width=0.151\textwidth]{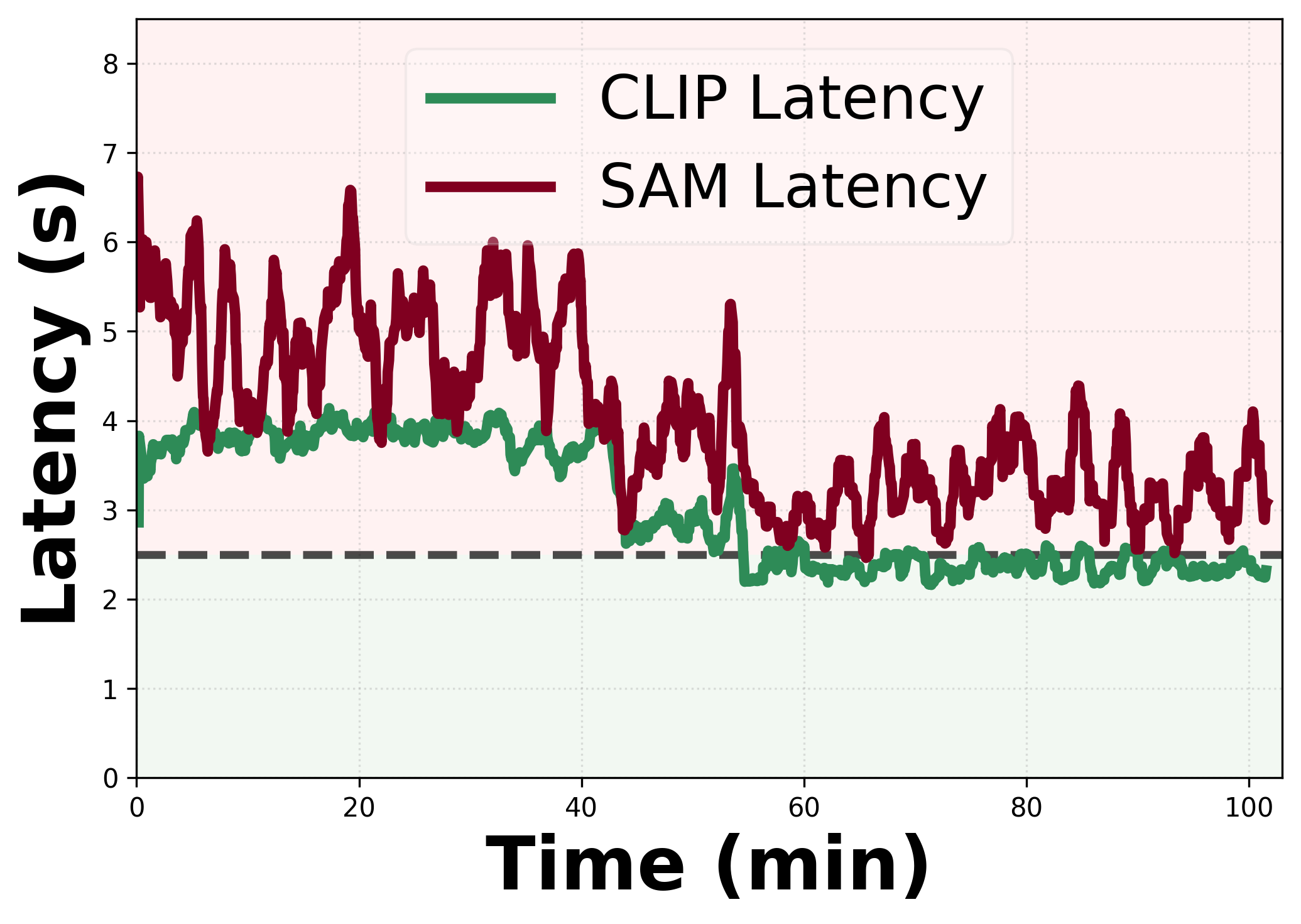}
    }\hfill
    \subfloat[Spot: Latency]{%
        \includegraphics[width=0.152\textwidth]{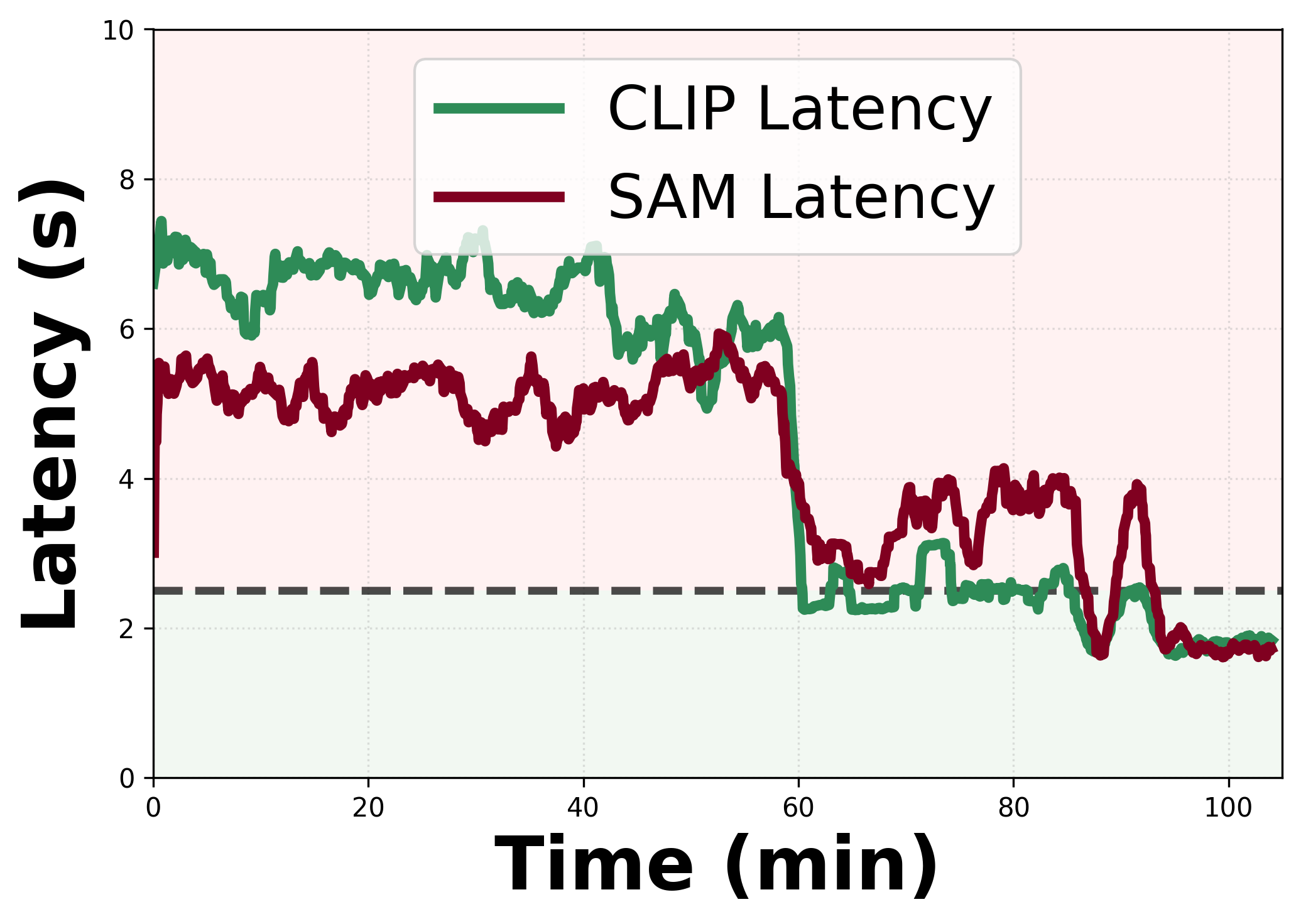}
    }

    \caption{Four-device setup results of the on-policy RL auction scheduler. FPS (top) and latency (bottom) for Husky, Jackal, and Spot show that adding an additional device preserves stable throughput and bounded latency, demonstrating scalable coordination under increased system heterogeneity.}
    \label{fig:fourth_results}
\end{figure}

After adding Linux device, the mean reward curves (Fig.~\ref{fig:spot_jackal_husky_reward_fourth}) show steady improvement and saturation after approximately 40–45 update steps, indicating stable on-policy MAPPO convergence as system scale increases. Following convergence, the FPS and latency results in Fig.~\ref{fig:fourth_results} show that the learned policy effectively leverages the extra execution capacity, maintaining stable throughput and bounded latency across all robots. Husky achieves consistently higher FPS with low deadline violations because its local compute capability is already strong, reducing the marginal benefit of offloading relative to Jackal and Spot. In contrast, the added executor provides clearer performance gains for the more resource-constrained platforms, demonstrating that the policy adapts to heterogeneous devices and scales gracefully with additional resources.

We evaluate the scalability and robustness of our RL-based task scheduler by varying both the number of devices and the workload mix (Table~\ref{tab:scalability_hosts_executor}). In the \textbf{3-device setup}, all three robots act as \emph{hosts}---each publishes workloads and participates in bidding/executing tasks---yielding success rates of \textbf{54.0\%} (Husky), \textbf{41.5\%} (Jackal), and \textbf{33.2\%} (Spot) (fig.~\ref{fig:success_rate}). We then extend this to a \textbf{4-device setup} by adding a \emph{Linux executor-only} node that does not publish workloads but can accept offloaded stages when it wins the bid. With this extra execution capacity, the success rates increase to \textbf{58.32\%} (Husky), \textbf{45.46\%} (Jackal), and \textbf{39.40\%} (Spot), showing that the framework can incorporate new devices without architectural changes and leverage them to reduce deadline violations.

We next consider a \textbf{1-device failure scenario} in which Spot is offline and only Husky and Jackal remain as hosts. Even after removing one robot, the system maintains competitive success rates of \textbf{50.30\%} (Husky) and \textbf{41.08\%} (Jackal). Relative to the 3-device setup, the performance drop is modest for Husky (\textbf{-3.70} points) and negligible for Jackal (\textbf{-0.42} points), indicating that the learned policy and auction mechanism degrade gracefully when resources shrink; the remaining hosts can still rebalance workloads and sustain near-nominal performance without retraining.

\begin{table}[b]
\caption{Scalability, fault-tolerance, and workload-stress evaluation}
\label{tab:scalability_hosts_executor}
\begin{tabular}{llll}
\hline
\textbf{Settings}                                                                                          & \textbf{Robot} & \textbf{Method}  & \textbf{\begin{tabular}[c]{@{}l@{}}Success \\ Rate (\%)\end{tabular}} \\ \hline
\multirow{3}{*}{\begin{tabular}[c]{@{}l@{}}4 devices: 3 hosts +\\ 1 executor-only \\ (Linux)\end{tabular}} & Husky          & RL Policy (ours) & 58.32                                                                 \\
                                                                                                           & Jackal         & RL Policy (ours) & 45.46                                                                 \\
                                                                                                           & Spot           & RL Policy (ours) & 39.40                                                                 \\ \hline
\multirow{3}{*}{\begin{tabular}[c]{@{}l@{}}2 devices: 2 hosts\\ (Spot offline)\end{tabular}}               & Husky          & RL Policy (ours) & 50.30                                                                 \\
                                                                                                           & Jackal         & RL Policy (ours) & 41.08                                                                 \\
                                                                                                           & Spot           & ---              & off                                                                   \\ \hline
\multirow{3}{*}{\begin{tabular}[c]{@{}l@{}}3 devices + one \\ extra task (yolo)\end{tabular}}              & Husky          & RL Policy (ours) & 49.1                                                                  \\
                                                                                                           & Jackal         & RL Policy (ours) & 39.8                                                                  \\
                                                                                                           & Spot           & RL Policy (ours) & 31.08                                                                
\end{tabular}
\end{table}

Finally, we stress-test the scheduler with a \textbf{3-device + extra task} setting in which we add the standard six-stage workload (3~SAM stages + 3~CLIP stages) with an additional object-detection stage based on YOLOv8-L~\cite{yolo}, a large-variant detector with tens of millions of parameters and substantially higher per-frame FLOPs). This significantly increases per-frame compute demand, yet the RL policy still achieves success rates of \textbf{49.1\%} (Husky), \textbf{39.8\%} (Jackal), and \textbf{31.08\%} (Spot). Compared to the original 3-device configuration, the degradation is moderate, especially for the more resource-constrained Spot, demonstrating that the learned RL auction policy remains effective even under heavier multi-model workloads and can absorb additional deep models without collapsing the overall latency/FPS success rate.

\section{Conclusion \& Future Work}

We present \names, a distributed and resource-aware framework for collaborative execution of large DNN and VLM inference workloads in multi-robot systems operating under stringent energy, network, and real-time constraints. Motivated by the limitations of centralized edge–cloud solutions in disaster-response and infrastructure-less environments, \namef enables autonomous robotic agents to cooperatively balance perception workloads based on dynamic resource availability and application-level performance requirements. By integrating a hybrid offline–online reinforcement learning strategy, \namef reduces the interaction cost typically associated with multi-agent learning while retaining adaptability to changing workloads, team compositions, and environmental conditions. Extensive real-world evaluations on heterogeneous robotic platforms demonstrate that \namef improves inference throughput, energy efficiency, and resilience compared to auction-based and learning-based baselines, while consistently meeting frame-rate and deadline constraints. 

Future work spans four key directions: (1) jointly optimizing perception, navigation, and communication for end-to-end mission efficiency; (2) integrating richer uncertainty modeling, such as probabilistic forecasts and confidence-aware VLM outputs, for robust scheduling under partial observability; (3) scaling to larger, intermittently connected teams via hierarchical coordination to minimize communication overhead; and (4) incorporating hardware-aware model compression, adaptive VLM partitioning, and continual learning to enhance long-term autonomy in highly dynamic environments.

\section{Acknowledgement}
This work has been partially supported by NSF CNS EAGER Grant \#2233879, ONR Grant \#N00014-23-1-2119, U.S. Army Grant \#W911NF2120076, U.S. Army Grant \#W911NF2410367, NSF CAREER Award \#1750936, and NSF REU Site Grant \#2050999.

\bibliographystyle{IEEEtran} 
\bibliography{references}

\end{document}